\documentclass[11pt]{article}
\usepackage[final]{acl}
\usepackage{times}
\usepackage{latexsym}
\usepackage[T1]{fontenc}
\usepackage[utf8]{inputenc}
\usepackage{microtype}
\usepackage{inconsolata}
\usepackage{amsmath,amssymb,amsthm,booktabs,array,tabularx}
\usepackage{colortbl}
\usepackage{enumitem}
\usepackage{graphicx}
\usepackage{placeins}
\usepackage[nameinlink,capitalise,noabbrev]{cleveref}


\definecolor{resultgreenbg}{RGB}{234,244,237}
\definecolor{resultgreentext}{RGB}{42,105,66}
\newcommand{\keyresult}[1]{\cellcolor{resultgreenbg}\textbf{#1}}
\renewcommand{\arraystretch}{1.04}
\newcommand{\method}{\textnormal{\textsc{FLCVA}}}
\newcommand{\answerop}{\mathcal{A}}
\newcommand{\ms}{M_s^{\answerop}}
\newcommand{\HeadlineTitle}{A Full-Ledger Counterfactual Audit of Adaptive Reasoning}

\title{\HeadlineTitle}
\author{Xiaopeng Chu \quad Wenzheng Kong \quad Wei Liu \quad Wenyi Zhang\thanks{Corresponding author.} \\
  University of Science and Technology of China \\
  Hefei 230027, China \\
  \texttt{\{cxphhh,wenzhengk,weiliu9416\}@mail.ustc.edu.cn} \\
  \texttt{wenyizha@ustc.edu.cn}}
\date{}

\begin{document}
\maketitle

\begin{abstract}
Adaptive reasoning must decide whether to answer now or spend more compute.
Predicting that the current answer is wrong does not establish whether
further reasoning will improve it.  We introduce the Full-Ledger
Counterfactual Value Audit (\method{}), which compares an immediate answer
with four fixed-length continuations from the same frozen prefix under one
answer procedure and accounts for the cost of every signal used by a policy.
On GSM8K and MATH-500, a random 256-token continuation recovers a wrong answer
in 26.6\% of Qwen3-8B cases and 18.9\% of DeepSeek-R1-Distill-Qwen-7B cases,
while damaging a correct answer in 1.4\% and 1.9\%.  Correctness labels from independently sampled
continuations predict a separate continuation's outcome in both primary
models.  In offline replay, using these labels to allocate compute improves
performance on DeepSeek; the Qwen3 gain remains inconclusive.  Under full
cost accounting, the tested lightweight, myopic controllers mainly recognize
current errors and add little continuation-specific value beyond that
shortcut.  A Llama-8B replication and action/cost ablations examine the
robustness of these findings.  \method{} separates
measuring local continuation value, predicting it from observable signals,
and using those predictions to allocate computation.

\end{abstract}

\section{Introduction}
\label{sec:introduction}

Longer reasoning can improve language-model accuracy, but it is not
automatically the best use of a limited inference budget.  Chain-of-thought
and zero-shot reasoning prompts elicit explicit reasoning, while
self-consistency, repeated sampling, verifier reranking, and search offer
several ways to spend additional compute
~\citep{wei2022chain,kojima2022zeroshot,wang2023selfconsistency,
brown2024monkeys,cobbe2021gsm8k,yao2023tot}.
Revision can improve an answer~\citep{madaan2023selfrefine}, but without
reliable feedback it can also reduce accuracy~\citep{huang2024selfcorrect}.
Test-time scaling must therefore decide not only whether more computation can
help, but \emph{where} and \emph{when} to spend it
~\citep{snell2024scaling,muennighoff2025s1}.

Adaptive methods use learned halting policies or signals such as confidence,
agreement, entropy, and predicted difficulty to halt, sample, branch, or route
computation
~\citep{graves2016act,banino2021pondernet,schuster2022calm,
aggarwal2023adaptive,damani2025hard,pan2025apr,scalena2025eager,
yang2025dynamicexit}.
These signals can identify an answer that is likely wrong
~\citep{kadavath2022know,kuhn2023semanticuncertainty,
xiong2024uncertainty,lightman2023verify},
but that is not the same as identifying whether further reasoning will repair
it.  Likewise, a stable answer may be correct and finished, wrong but
recoverable, or correct and vulnerable to revision.  The decision-relevant
object is a checkpoint-level counterfactual: from an identical frozen prefix,
what changes if the model answers now rather than receiving one more
predeclared block of reasoning?

\begin{figure*}[t]
\centering
\includegraphics[width=0.87\linewidth,trim=0 18bp 0 5bp,clip]{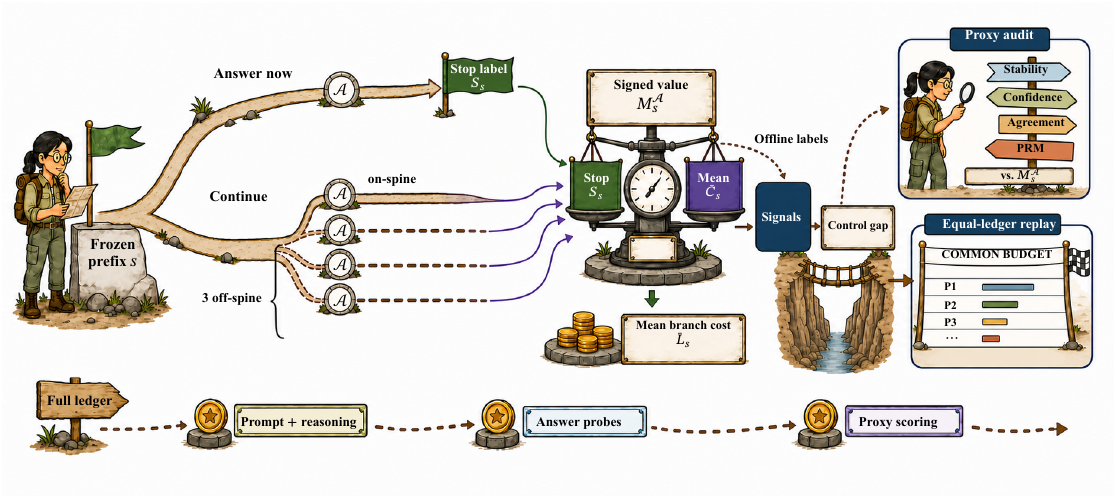}
\caption{\method{} separates measurement from control.  From one frozen
prefix, the same answer procedure evaluates answering now, continuing along
the stored trajectory for another 256 tokens, and taking three separately
sampled continuations of the same length.  Correctness change divided by
charged cost defines signed local value.  Continuation correctness labels
are available only offline; replay policies use inference-time signals and
pay for every observation they require.}

\label{fig:protocol-overview}
\end{figure*}

\begin{figure}[t]
\centering
\includegraphics[width=\columnwidth]{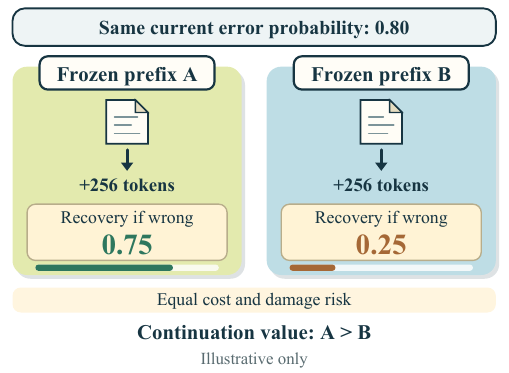}
\caption{Two prefixes can have the same current-error probability but
different recovery probabilities.  With equal continuation costs and equal
damage risk for correct answers, their expected continuation values differ.
Probabilities are illustrative, not experimental estimates.}
\label{fig:two-prefix-example}
\end{figure}

\method{}\footnote{Code and evaluation scripts are available at
\url{https://github.com/cxp-ustc/flcva}.}
compares answering now with continuing from the same \emph{frozen prefix}:
the prompt and all reasoning tokens generated so far.  The audit grades each
outcome offline; a deployed policy must act without those future correctness
labels.  \Cref{fig:protocol-overview} shows the protocol, and
\Cref{fig:two-prefix-example} illustrates why current error probability alone
cannot determine continuation value.

We separate three questions.  \textbf{Measurement:} how much does a specified
continuation change correctness per unit cost?  \textbf{Prediction:} do
observable signals distinguish recoverable errors and vulnerable correct
answers?  \textbf{Allocation:} do these predictions improve accuracy under a
shared compute budget after paying for the signals?  For allocation,
comparison with a current-wrongness baseline tests the additional value of
predicting which errors can be repaired.

Our contributions are:
\begin{itemize}[leftmargin=*,topsep=2pt,itemsep=1.5pt,parsep=0pt]
  \item \textbf{A paired audit of continuation value.}  We compare answering
  now with fixed-length continuations from the same prefix under a fixed
  answer procedure, measuring signed correctness change per expected logical
  token.
  \item \textbf{A full-ledger proxy audit.}  We evaluate observable signals
  against this target and charge every observation required by a policy,
  including the cost of obtaining its signals.
  \item \textbf{A measurement-to-control decomposition.}  We distinguish
  offline diagnostics using future correctness labels from deployable
  policies.  Scoring with independently sampled branches avoids leakage from
  the evaluated continuation.  Comparison with a current-wrongness baseline
  tests allocation gains beyond error recognition.
  \item \textbf{Cross-model evidence and intervention robustness.}  Across
  two primary models and a Llama-8B replication, continuation can repair errors
  and damage correct answers.  Persistence provides information beyond
  current correctness, but the tested lightweight, myopic controllers add
  little beyond error recognition.  We examine sensitivity to continuation
  length, answer cues, parsers, and logical and physical cost accounting.
\end{itemize}

\section{Related Work}
\label{sec:related}

\paragraph{Adaptive inference and value of computation.}
Metareasoning and anytime algorithms frame computation as an action whose
value depends on the current decision state
~\citep{horvitz1989reflection,zilberstein1996anytime,hay2012selecting}.
Modern test-time reasoning instantiates this idea through adaptive sampling,
learned routing, branch allocation, confidence-based stopping, and budget
forcing
~\citep{aggarwal2023adaptive,damani2025hard,pan2025apr,scalena2025eager,muennighoff2025s1}.
\citet{wu2025inferencescaling} compare model sizes and inference strategies
under compute budgets, showing how the inference procedure shapes the
accuracy--cost trade-off.
\method{} instantiates value of computation as paired measurement of local
stop--continue value for a realized language-model prefix, complementing work
whose primary contribution is a controller.

\paragraph{Confidence and process verification.}
Self-evaluation, semantic uncertainty, outcome verifiers, PRMs, and
process-advantage verifiers estimate answer reliability or intermediate
progress
~\citep{kadavath2022know,kuhn2023semanticuncertainty,
xiong2024uncertainty,uesato2022processoutcome,lightman2023verify,
wang2023mathshepherd,zhang2025prm,setlur2024progress}.
SCoRe uses multi-turn reinforcement learning to improve a model's ability to
correct its own errors~\citep{kumar2025score}.
We test how well these allocation signals predict the correctness change
from a specified continuation.  The public PRM serves as an observable
proxy, and external endpoint grading defines the audit target.

\paragraph{Closest adaptive-reasoning systems.}
Studies of overthinking identify redundant reasoning that adds cost with
little accuracy benefit~\citep{chen2025overthinking}.
TRACE combines temporal consistency and confidence; UAB allocates sampling
from question uncertainty; PoLR and Prefix Consistency use truncated prefixes
or regenerated continuations; ThinkBooster unifies scaling strategies and
cost views; and PUMA combines semantic redundancy with answer verification
~\citep{li2026trace,nguyen2026uab,jindal2026polr,iwase2026prefix,smirnov2026thinkbooster,min2026puma}.
These works primarily propose controllers, reliability signals, or evaluation
frameworks.  \method{} instead makes the paired stop--continue contrast the
estimand and jointly audits proxy information and charged cost.
\Cref{tab:related-comparison} makes this difference explicit.

\begin{table*}[t]
\centering
\fontsize{8.0pt}{10pt}\selectfont
\setlength{\tabcolsep}{3.5pt}
\renewcommand{\arraystretch}{0.98}
\begin{tabular}{@{}p{0.125\textwidth}p{0.205\textwidth}p{0.270\textwidth}
p{0.145\textwidth}p{0.185\textwidth}@{}}
\toprule
Work & Primary decision object & Signal or intervention &
Paired same-prefix signed target & Cost role \\
\midrule
APR & Serial/parallel scheduling & Learned spawn-and-join policy &
No & Compute and latency \\
EAGer & Branch/sample allocation & Token-entropy branching &
No & Token efficiency \\
TRACE & Early exit & Temporal consistency and confidence &
No & Accuracy--token tradeoff \\
UAB & Per-question sample count & Initial-sample vote entropy &
No & Fixed sampling budget \\
PoLR & Prefix-cluster expansion & Consensus among short prefixes &
No & Tokens and latency \\
Prefix Consistency & Candidate weighting & Regeneration from truncated prefixes &
Partial & Voting cost \\
ThinkBooster & Test-time scaling strategy & Modular scorers and strategies &
Varies & Quality--cost evaluation \\
PUMA & Early exit & Redundancy plus answer verification &
No & Token reduction \\
\method{} & Measurement and allocation audit &
Answer-now versus four fixed continuations &
\textbf{Yes} & Full charged ledger \\
\bottomrule
\end{tabular}
\caption{Closest-system comparison.  A paired target contrasts answer-now and
continuation from the same frozen prefix; ``Partial'' denotes related
truncated-prefix signals without this signed contrast.}
\label{tab:related-comparison}
\end{table*}

\section{The \method{} Measurement}
\label{sec:method}

\subsection{Signed continuation value}

At each checkpoint, we keep the prompt and existing reasoning tokens fixed.
We obtain an answer immediately and another answer after each of four
continuations.  All answers use the same generation, parsing, and grading
procedure, called the \emph{answer operator}.  We subtract immediate-answer
correctness from average continuation correctness and divide by average
continuation cost, including the answer tokens.  The resulting value is
positive for observed repair, negative for observed damage, and zero when
none of the sampled continuations changes correctness.

Let \(s\) be a frozen reasoning prefix and let \(\answerop\) denote the fixed
answer suffix, greedy decoding rule, parser, normalizer, and exact-match
grader.  Answering immediately gives correctness \(S_s\in\{0,1\}\).
Continuation branch \(k\) gives correctness \(C_{s,k}\) after a fixed
reasoning action and incurs logical cost \(L_{s,k}>0\), including its answer
probe.  For \(K=4\), we define
\begin{equation}
  \ms =
  1000\,
  \frac{K^{-1}\sum_{k=1}^{K} C_{s,k}-S_s}
       {K^{-1}\sum_{k=1}^{K} L_{s,k}}.
  \label{eq:value}
\end{equation}
The denominator normalizes by logical token cost.  Multiplication by 1,000
expresses the result as expected correctness change \emph{per 1,000 logical
tokens}, making small per-token values easier to read.  This common positive
scaling leaves the signs and rankings in Eqs.~(1)--(3) unchanged.

Equation~\ref{eq:value} is a ratio of average correctness change to average
cost, rather than an average of branch-specific ratios.  Its target is
conditional on the answer operator, sampling scheme, and continuation length.
Recovery is possible only when \(S_s=0\), and damage only when \(S_s=1\).

Let \(p_s=\frac14\sum_k C_{s,k}\).  We report two distinct estimands.  For a
currently wrong answer, \(p_s\) estimates single-draw recovery probability,
and \(\mathbf 1[p_s>0]\) detects recovery in any of four sampled branches.
For a currently correct answer, \(1-p_s\) estimates single-draw damage
probability, and \(\mathbf 1[p_s<1]\) detects any observed damage.  Four-draw
detection measures whether the sampled branches contain at least one
recovery or damage event.

\subsection{Protocol and evaluation split}

We evaluate Qwen3-8B and DeepSeek-R1-Distill-Qwen-7B
~\citep{yang2025qwen3,deepseekai2025r1} on all 1,319 GSM8K test problems and
all 500 MATH-500 problems~\citep{cobbe2021gsm8k,hendrycks2021math}.  Each
model--problem pair has four independently sampled reasoning trajectories,
called \emph{spines}, capped at 4,096 tokens and frozen every 256 tokens.
At every nonterminal checkpoint we apply answer-now and four 256-token
continuations.  Branch 1 reuses the next stored trajectory segment (the
\emph{on-spine} branch); branches 2--4 are sampled separately from the same
prefix (\emph{off-spine} branches).  The same greedy answer operator grades
all endpoints.

We deterministically partition problem IDs, assigning 80\% of problems to
analysis and 20\% to confirmation.  We use analysis problems to choose each
proxy's scoring direction, calibrate predictions, and fit controller
parameters.  Confirmation continuation outcomes are used only for held-out evaluation.
Unless stated otherwise, intervals use 2,000 bootstrap draws of complete
problems, retaining their models, spines, checkpoints, and branches.  The
primary study contains 14,552 spines and 218,280 eligible checkpoints.
Details of sampling, parsing, missing data, and bootstrapping are in
\Cref{app:experimental-details,app:proxy-detail}.

\subsection{Identification and uncertainty}

The paired comparison holds the prompt and frozen reasoning tokens
identical across the answer-now arm and all continuations.  The arms differ
in the action taken after checkpoint \(s\).  Branch 1 reuses the stored spine
segment while retaining its full reasoning-plus-probe cost in \(L_{s,1}\).
This saves physical generation while preserving the specified intervention
and its logical cost.  Later checkpoints follow the original spine,
independently of the outcomes of off-spine samples.

We estimate uncertainty by resampling benchmark problems.  Each problem
contributes multiple models, spines, depths, and correlated outcomes, so each
bootstrap draw retains all records nested within the selected problem IDs.
Point estimates weight eligible checkpoints in the reported population;
intervals preserve within-problem dependence and quantify uncertainty across
problems conditional on the stored trajectories.

\subsection{Proxies and cost}

We audit recent-answer persistence, answer confidence and its negative-log
form (ANLL), four-spine modal agreement, minimum step-level process reward
model (PRM) score, and top-20 approximate entropy.  Persistence is the fraction
of up to four recent stop answers matching the current answer.  Confidence
and ANLL satisfy \(\mathrm{confidence}=\exp(-\mathrm{ANLL})\), so they encode
the same signal with reversed order; we summarize them as one signal family.
The controllers use both input features.

The primary prediction metric is AUROC for detecting any recovery among
currently wrong checkpoints and any damage among currently correct ones.
These labels indicate whether at least one of four branches changes
correctness; they differ from the single-draw probabilities above.  Average
precision with prevalence baselines and calibration assess rare events and
probability quality.  Matched and conditional diagnostics test information
beyond model, dataset, depth, difficulty, and current correctness; they are
grouped in \Cref{app:proxy-detail}.  Difficulty is a label-aware diagnostic
stratifier and is never exposed to a deployable policy.

The primary ledger counts all logical tokens needed to expose a prefix,
obtain an answer, advance a spine, or score a policy-specific proxy,
including the observations required by PRM and parallel-agreement policies.
Logical tokens quantify intervention cost.  To examine hardware-related
costs, we also replay policy orderings under estimated FLOPs and calibrated
H20 seconds and report measured GPU time (\Cref{app:cost-sensitivity}).

\section{From Measurement to Allocation}
\label{sec:allocation}

A \emph{frontier} is a problem's current checkpoint during compute
allocation.  Replay begins with one frontier per confirmation problem and a
shared budget.  Each action advances one problem by the next stored 256-token
segment and reveals its next answer.  The four spine seeds define four
repetitions, with one spine per problem in each repetition and accuracy
measured over problems.  All policies are compared up to the same reachable
budget endpoint.  We measure budget-averaged accuracy using normalized area
under the accuracy--ledger curve (AULC).  For each full-system bootstrap
replicate, we resample problems, rerun allocation, and recompute the budget
endpoints.

For repetition \(r\), the reported score is
\[
  \operatorname{AULC}_r=
  \frac{1}{B_r-B_{0,r}}
  \int_{B_{0,r}}^{B_r}\operatorname{Acc}_r(b)\,db ,
\]
where \(B_{0,r}\) includes every observation required to expose the initial
frontier and \(B_r\) is the common endpoint.  Final accuracy is the same when
all stored transitions are exhausted; AULC measures how efficiently a policy
improves accuracy along the way.

We compare three levels of allocation.  \emph{Round-robin} provides a
reference for measuring adaptive allocation gains.  A \emph{correctness shortcut} predicts only
whether the current answer is wrong and combines that probability with
analysis-set dataset--depth mean recovery, damage, and cost.  Finally,
continuation-specific policies either factor expected value into conditional
recovery and damage probabilities (two-stage EV) or directly regress signed
\(\ms\) (Ridge-\(M\)).  Their inputs are single-spine proxies, depth, and
dataset.  Comparing with round-robin measures overall allocation gains;
comparing with the correctness shortcut measures gains beyond error
recognition.

Concretely, let \(q_s\) predict that the current answer is wrong, and let
\(r_s,d_s\) predict recovery and damage conditional on current correctness.
With analysis-set dataset--depth means \(\bar r_{d,t},\bar d_{d,t}\) and mean
branch cost \(\bar L_{d,t}\), the priorities are
\begin{align}
\widehat V_s^{\mathrm{shortcut}}
  &=1000\frac{q_s\bar r_{d,t}-(1-q_s)\bar d_{d,t}}{\bar L_{d,t}}, \\
\widehat V_s^{\mathrm{two}}
  &=1000\frac{q_sr_s-(1-q_s)d_s}{\bar L_{d,t}}.
\end{align}
The second uses conditional recovery and damage predictions to distinguish
repairable errors and vulnerable correct answers beyond current error probability.
Both priorities are \emph{myopic}: they score the next action without valuing
the later checkpoints it makes available.  Our primary allocation endpoint
is the percentage change in normalized AULC relative to the correctness
shortcut, with paired 95\% full-system bootstrap intervals.

Offline diagnostics use continuation correctness labels to assess allocation
opportunity.  The mixed four-branch value includes the realized replay
successor, branch 1, and therefore incorporates hindsight.  The off-spine
score ranks the on-spine action using branches 2--4, separating the scoring
labels from the realized successor.  These labels are used only for offline
evaluation.  Exact dynamic programming provides a multi-step upper bound
using known future outcomes on deterministically selected four-problem
subsets.  Details of policy inputs, alternative replay units,
leave-one-branch-out validation, and the DP setup are in
\Cref{app:replay-controls}.

\begin{table*}[t]
\centering
\fontsize{8pt}{10pt}\selectfont
\setlength{\tabcolsep}{3pt}
\renewcommand{\arraystretch}{0.98}
\begin{tabular}{@{}p{0.14\textwidth}p{0.20\textwidth}p{0.15\textwidth}p{0.15\textwidth}p{0.30\textwidth}@{}}
\toprule
Policy & Required observation & Extra generation & Extra scoring & Replay charge \\
\midrule
Round-robin / fixed & One current spine & None & None & Prompt, reasoning, answer probe \\
TRACE / ANLL / entropy & Current/prior single-spine probes & None & None & Single-spine ledger \\
PRM & Current single-spine prefix & None & PRM prefill & Ledger plus every PRM call \\
Cross-spine agreement & Four equal-depth answers & Three synchronized spines & None & Four reasoning/probe paths \\
Ridge / two-stage & Single-spine proxies and context & None & Optional PRM & All observed inputs \\
Mixed/off-spine/DP & Continuation correctness labels & Stored branches/futures & Gold grading & Diagnostic only \\
\bottomrule
\end{tabular}
\caption{Policy observability and accounting.  Cross-spine agreement charges
all parallel trajectories used to obtain the signal.  Label-aware rows
provide offline allocation diagnostics.}
\label{tab:policy-accounting}
\end{table*}

The observability and charging boundary for every replay policy is summarized
in \Cref{tab:policy-accounting}.

\section{Results}
\label{sec:experiments}

\paragraph{Evaluation scope and primary endpoints.}
The primary confirmation split contains 372 problems evaluated by each
primary model.  Every observable controller is fitted only on the disjoint
analysis split.  The evidence follows three stages:
\textbf{measurement} characterizes signed continuation value and its recovery
and damage components (\S5.1);
\textbf{prediction} tests conditional recovery/damage AUROC (\S5.2);
and \textbf{allocation} tests relative normalized-AULC gain over the
correctness shortcut (\S5.3).  Opportunity diagnostics (\S5.4) and
robustness analyses (\S5.5) explain the scope of these results.

\subsection{Continuation frequently repairs errors, and sometimes damages answers}

\begin{table*}[t]
\centering
\footnotesize
\begin{tabular}{llrr}
\toprule
Model & Event & Single draw & Any of four \\
\midrule
Qwen3 & Recovery & \keyresult{26.6\% [22.5, 31.6]} & 35.7\% [30.3, 42.3] \\
Qwen3 & Damage & \keyresult{1.4\% [1.0, 1.8]} & 2.9\% [2.3, 3.6] \\
DeepSeek & Recovery & \keyresult{18.9\% [16.2, 22.2]} & 26.0\% [22.2, 30.2] \\
DeepSeek & Damage & \keyresult{1.9\% [1.5, 2.3]} & 3.6\% [3.0, 4.2] \\
\bottomrule
\end{tabular}
\caption{Single-draw transition rates and four-draw opportunity detection on
confirmation problems.  Brackets are 95\% problem-bootstrap intervals.}
\label{tab:branch-estimands}
\end{table*}

\begin{figure*}[t]
\centering
\includegraphics[width=0.6\linewidth]{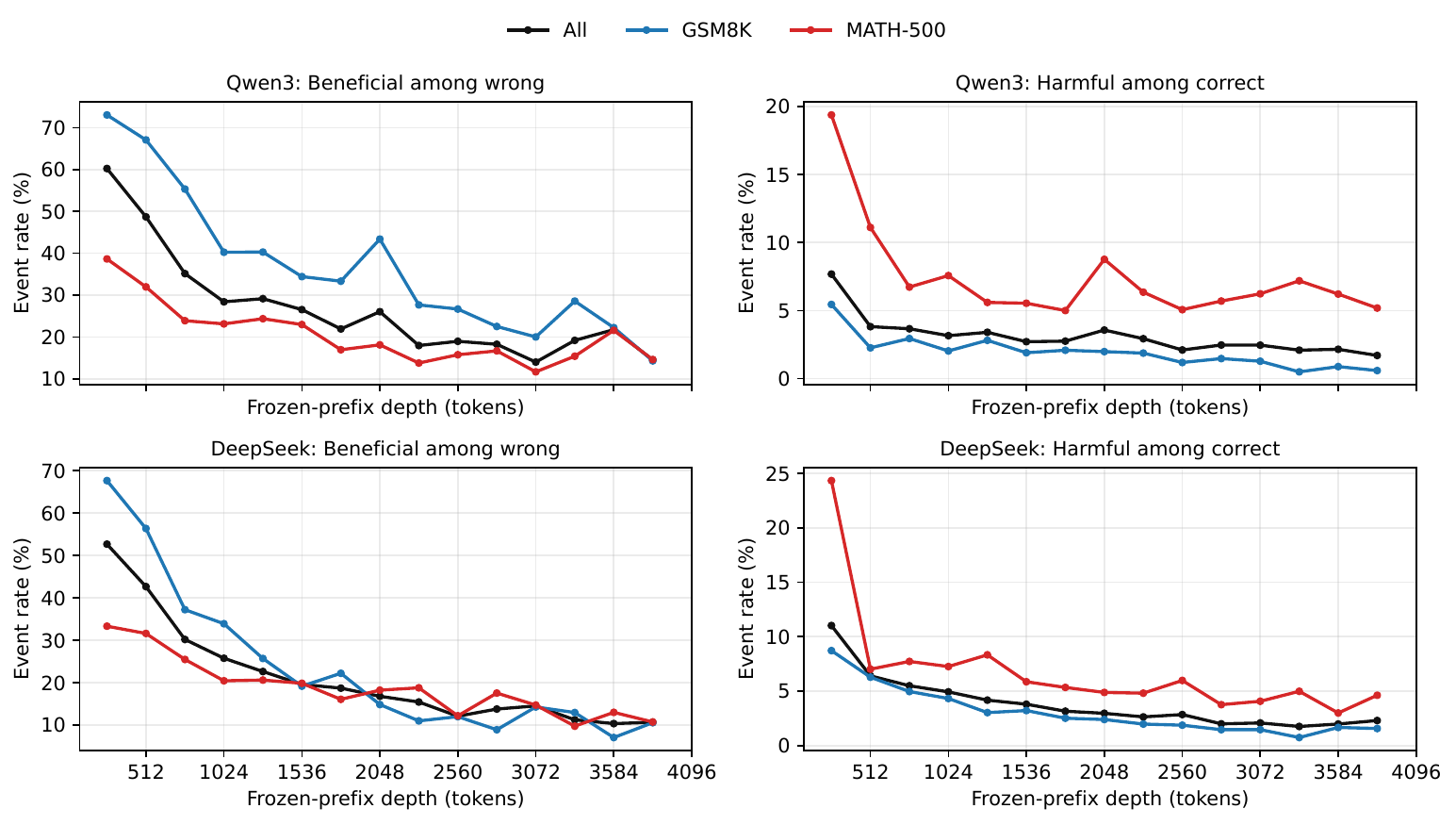}
\caption{Recovery and damage prevalence by prefix depth and dataset on
confirmation problems.  Recovery conditions on the current answer being
wrong; damage conditions on it being correct.}
\label{fig:rq1-depth}
\end{figure*}

For a uniformly selected continuation, recovery occurs in 26.6\% of Qwen3
wrong checkpoints and 18.9\% of DeepSeek wrong checkpoints
(\Cref{tab:branch-estimands}).  Damage is rarer but nonzero: 1.4\% and 1.9\%
among currently correct checkpoints.  With four sampled branches, at least
one recovery appears at 35.7\% and 26.0\% of wrong checkpoints, while at least
one damaging branch appears at 2.9\% and 3.6\% of correct checkpoints.  These
rates vary with dataset and depth (\Cref{fig:rq1-depth}); complete denominators
are in \Cref{app:intervention-robustness}.
Conditional on detecting an event, mean benefit magnitude is 2.865 and 2.794
value units, versus harm magnitudes of 1.817 and 2.000.  Four-draw harm is also
more frequent on MATH-500 (6.7\% and 6.0\%) than on GSM8K (1.8\% and 2.9\%).

Among currently correct checkpoints, greater persistence is associated with
less damage.  The high-minus-low persistence contrast in signed value is
\(+0.200\) for Qwen3 (95\% CI \([0.135,0.276]\)) and \(+0.277\) for
DeepSeek (\([0.201,0.361]\)).  Since value is nonpositive in this population,
positive contrasts mean that highly persistent answers are less vulnerable
to harmful revision.

\subsection{Proxies contain information without determining local value}

\begin{table}[t]
\centering
\footnotesize
\setlength{\tabcolsep}{3pt}
\begin{tabular}{lllr}
\toprule
Model & Target & \shortstack[l]{Best proxy\\(available $n$)} & \shortstack[r]{AUROC\\(95\% CI)} \\
\midrule
Qwen3 & Recovery & \shortstack[l]{Persistence\\(2,702)} & \shortstack[r]{0.658\\{[0.610, 0.706]}} \\
Qwen3 & Damage & \shortstack[l]{Agreement\\(18,634)} & \shortstack[r]{0.742\\{[0.699, 0.782]}} \\
DeepSeek & Recovery & \shortstack[l]{Persistence\\(3,665)} & \shortstack[r]{0.673\\{[0.632, 0.715]}} \\
DeepSeek & Damage & \shortstack[l]{Persistence\\(17,139)} & \shortstack[r]{0.716\\{[0.681, 0.747]}} \\
\bottomrule
\end{tabular}
\caption{Best confirmation proxy by AUROC for detecting any recovery among
wrong checkpoints or any damage among correct checkpoints over four
continuations.  Counts reflect proxy availability; intervals resample whole
problems.  Rare-event baselines and calibration appear in
\Cref{app:proxy-detail}.}
\label{tab:best-proxy-ci}
\end{table}

Persistence is the strongest recovery proxy in both models, with AUROC 0.658
and 0.673.  Modal agreement is strongest for Qwen3 damage (0.742), while
persistence is strongest for DeepSeek damage (0.716)
(\Cref{tab:best-proxy-ci}).  These AUROCs evaluate detection of recovery or
damage among four sampled continuations.

Persistence also retains information after accounting for current
correctness, dataset, depth, and difficulty: conditional AUROC is
0.632--0.692 and held-out log-loss reduction is 0.021--0.053 bits.  Thus it
provides information about recovery and damage beyond identifying current
errors.  Matched checkpoints with the same signal value can have different
outcomes.  Calibration, rare-event baselines, and matched-group diagnostics
in \Cref{app:proxy-detail} further characterize this variation.

\subsection{Observable adaptation mostly learns current wrongness}

\begin{table*}[t]
\centering
\footnotesize
\setlength{\tabcolsep}{2.5pt}
\begin{tabular}{llrrr}
\toprule
Model & Policy & Norm.\ AULC & Gain vs.\ RR & Increment vs.\ shortcut \\
\midrule
Qwen3 & Round-robin & 0.851 & +0.00\% [+0.00, +0.00] & -1.75\% [-2.72, -0.72] \\
Qwen3 & Correctness shortcut & 0.866 & \keyresult{+1.78\% [+0.73, +2.80]} & +0.00\% [+0.00, +0.00] \\
Qwen3 & Two-stage conditional EV & 0.860 & +1.08\% [+0.28, +1.91] & \keyresult{-0.69\% [-1.51, +0.10]} \\
Qwen3 & Direct Ridge-$M$ & 0.864 & +1.57\% [+0.65, +2.51] & \keyresult{-0.21\% [-0.52, +0.12]} \\
\addlinespace
DeepSeek & Round-robin & 0.813 & +0.00\% [+0.00, +0.00] & -1.53\% [-2.60, -0.44] \\
DeepSeek & Correctness shortcut & 0.825 & \keyresult{+1.55\% [+0.44, +2.67]} & +0.00\% [+0.00, +0.00] \\
DeepSeek & Two-stage conditional EV & 0.826 & +1.60\% [+0.44, +2.77] & \keyresult{+0.05\% [-0.96, +1.00]} \\
DeepSeek & Direct Ridge-$M$ & 0.831 & +2.26\% [+1.40, +3.17] & \keyresult{+0.70\% [+0.07, +1.35]} \\
\addlinespace
\bottomrule
\end{tabular}
\caption{Observable-controller hierarchy under the problem-level single-spine
logical ledger.  The final column measures allocation gains beyond the
correctness shortcut.  Brackets are paired 95\% full-system problem-bootstrap
intervals.}
\label{tab:controller-decomposition-replay}
\end{table*}

\begin{figure}[t]
\centering
\includegraphics[width=0.8\columnwidth]{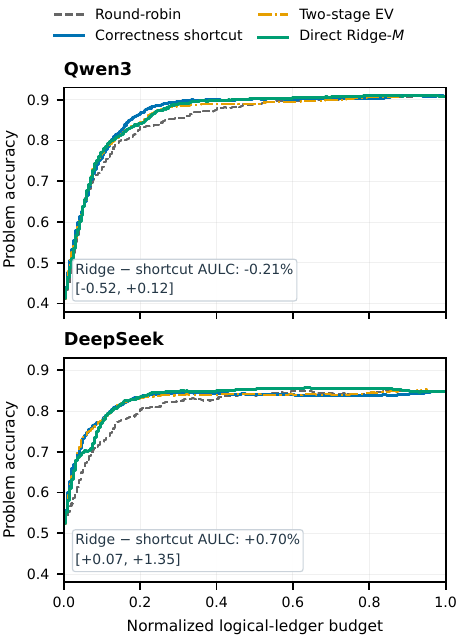}
\caption{Held-out controller accuracy under a shared logical-token budget, using one spine
per problem.  Curves average four spine-seed repetitions; annotations give
Ridge-\(M\)'s relative gain in normalized AULC over the correctness shortcut,
with full-system 95\% intervals.}
\label{fig:controller-hierarchy}
\end{figure}

The correctness shortcut already improves relative normalized AULC over
round-robin by \(+1.78\%\) on Qwen3 and \(+1.55\%\) on DeepSeek
(\Cref{tab:controller-decomposition-replay,fig:controller-hierarchy}).  Relative to this stronger
baseline, two-stage EV changes AULC by \(-0.69\%\)
(\([-1.51,0.10]\)) and \(+0.05\%\) (\([-0.96,1.00]\)); direct Ridge-\(M\)
changes it by \(-0.21\%\) (\([-0.52,0.12]\)) and \(+0.70\%\)
(\([0.07,1.35]\)).  Thus Qwen3 shows no reliable continuation-specific
increment, while DeepSeek shows a modest one.  TRACE, PRM, ANLL, and entropy
also have point estimates at or below the shortcut.  Conditional recovery and
damage AUROCs are 0.767--0.799, with limited allocation gains over the
shortcut.  AUROC evaluates ranking within each outcome population, whereas
replay combines recovery and damage predictions with costs to repeatedly
rank 372 changing frontiers.  The following controls examine how signal
costs and one-step scheduling affect allocation.

\begin{table}[t]
\centering
\footnotesize
\setlength{\tabcolsep}{3pt}
\begin{tabular}{lrrr}
\toprule
Model & \shortstack{Current\\wrong} & Recovery & Damage \\
\midrule
Qwen3 & 0.882 & 0.774 & 0.799 \\
DeepSeek & 0.843 & 0.767 & 0.772 \\
\bottomrule
\end{tabular}
\caption{Held-out prediction by the two-stage model.  All components are fitted
on analysis problems, and replay uses observable features from confirmation
problems.}
\label{tab:controller-decomposition-predictive}
\end{table}

\Cref{tab:controller-decomposition-predictive} separates current-wrongness
prediction from conditional recovery and damage prediction, clarifying what
each controller can learn from observable features.

\subsection{Independent branches reveal allocation opportunity}

\begin{table}[t]
\centering
\footnotesize
\setlength{\tabcolsep}{2.8pt}
\begin{tabular}{lcc}
\toprule
Model & \shortstack{Off-spine\\gain vs.\ RR} &
\shortstack{Exact-DP\\gain vs.\ RR} \\
\midrule
Qwen3 & \shortstack{\(+1.44\%\)\\\([-0.48,\,3.37]\)} &
\keyresult{\shortstack{\(+7.08\%\)\\\([+5.31,\,+9.03]\)}} \\
DeepSeek & \keyresult{\shortstack{\(+5.41\%\)\\\([+3.61,\,+7.29]\)}} &
\keyresult{\shortstack{\(+11.55\%\)\\\([+8.19,\,+15.11]\)}} \\
\bottomrule
\end{tabular}
\caption{Offline allocation opportunity under the logical-token ledger.
Off-spine ordering uses labels from branches sampled separately from the
realized successor; exact DP gives a hindsight bound on four-problem subsets.}
\label{tab:arr-opportunity-gap}
\end{table}

In the offline diagnostic, correctness labels from three off-spine branches
predict the outcome of the held-out on-spine continuation in both primary
models.  Using this estimate to order replay yields \(+1.44\%\) relative
normalized AULC for Qwen3, with an interval crossing zero, and \(+5.41\%\)
for DeepSeek, with a positive interval (\Cref{tab:arr-opportunity-gap}).
Direct cross-branch prediction of the realized successor reaches
0.859--0.948 AUROC across events and models
(\Cref{tab:leave-one-branch-out}).

\paragraph{Effects of signal cost and scheduling.}
Two controlled comparisons examine signal costs and scheduling.  First, we
fix the action order based on agreement and charge the extra trajectories
needed to obtain that signal.  DeepSeek normalized AULC falls from 0.819 to
0.672 (\Cref{tab:agreement-accounting}), isolating the effect of including
signal costs.

Second, we compare greedy allocation by true next-step correctness change
per token with exact multi-step allocation on the same 64 non-overlapping
four-problem subsets per model.  Both schedules use future outcomes.
Normalized AULC is 0.820 versus 0.902 for Qwen3 and 0.791 versus 0.854 for
DeepSeek (\Cref{tab:nonmyopic-dp}).  This hindsight comparison shows the
benefit of planning beyond immediate value within these subsets.

Together, these controls motivate charging every signal and considering the
value of subsequent states.  The relative contributions of signal quality,
calibration, and planning horizon to the observed controller gap remain an
open question.

\begin{table}[t]
\centering
\footnotesize
\setlength{\tabcolsep}{2.5pt}
\begin{tabular}{llrrr}
\toprule
Model & Length & Rec.\% & Dmg.\% & \(\Delta M\) \\
\midrule
Qwen3 & 128 & 19.0 & 1.6 & +0.016 \\
Qwen3 & 256 & 26.6 & 1.5 & 0.000 \\
Qwen3 & 512 & 36.0 & 1.6 & -0.031 \\
Qwen3 & Adaptive & 26.8 & 1.6 & 0.000 \\
\addlinespace
DeepSeek & 128 & 14.3 & 1.7 & +0.021 \\
DeepSeek & 256 & 19.6 & 1.9 & 0.000 \\
DeepSeek & 512 & 25.5 & 2.0 & -0.023 \\
DeepSeek & Adaptive & 14.9 & 1.9 & \keyresult{+0.023} \\
\bottomrule
\end{tabular}
\caption{Continuation-length sensitivity on the 200-problem confirmation
ablation.  Recovery and damage are single-draw rates; \(\Delta M\) is the change
in value per 1,000 logical tokens relative to fixed 256.  Full intervals
appear in \Cref{tab:block-length-events-full}.}
\label{tab:block-length-events}
\end{table}

\subsection{Robustness and external replication}

\paragraph{Continuation length.}
In a paired 200-problem confirmation ablation, recovery rises with
128/256/512-token continuations: 19.0/26.6/36.0\% for Qwen3 and
14.3/19.6/25.5\% for DeepSeek.  Paired differences in damage rates are not
distinguishable from zero.  For each candidate length \(\ell\), a ridge model
fitted on analysis problems predicts signed change per 1,000 logical tokens
from dataset and checkpoint depth.  On confirmation problems, the frozen
selector chooses \(\arg\max_\ell \widehat V_\ell\), breaking ties toward the
shorter action.  Within each model, the selector is a local rule based only on dataset
and depth.  \Cref{tab:block-length-events} reports event rates and
full-sample proxy rankings.  On DeepSeek, the selector improves local value
by \(+0.023\) (\([0.007,0.039]\)) correctness-change units per 1,000 logical
tokens, largely by selecting the cheaper 128-token action.
Mean signed value per 1,000 logical tokens falls with length, from
0.137/0.113 at 128 tokens to 0.090/0.068 at 512 for Qwen3/DeepSeek.
The fixed 256-token action lies between shorter, cheaper continuations and
longer ones that repair more errors at lower value per token.  Persistence remains the leading recovery proxy at all three
lengths in the full sample, although its ranking is less stable in the
512-token half-sample analysis.

\paragraph{Answer procedure and grading.}
A similar, concise answer cue produces parseable answers in 99.8--99.9\% of
cases and preserves 89.8--91.0\% of recovery labels and 96.5--97.0\% of damage
labels.  A more directive cue substantially changes the audit target.
Across sampled groups, independent rational/symbolic regrading agrees with
95.5--100\% of primary transition labels among cases it can resolve.

\paragraph{Cost accounting.}
After charging their costs, PRM and parallel-agreement policies remain below
single-spine baselines under logical tokens, estimated FLOPs, and calibrated
H20 seconds.  Across these ledgers, PRM trails round-robin by 33.5--48.4\% on
Qwen3 and 25.2--33.9\% on DeepSeek, showing a consistent cost disadvantage
across all three measures.

\paragraph{Replication on Llama.}
The full protocol on DeepSeek-R1-Distill-Llama-8B yields 10.2\% single-draw
recovery and 0.9\% damage.  Greater persistence is again associated with less
damage (\Cref{tab:llama-estimands}).  Using predictors fitted on the two
primary models without refitting, the correctness shortcut gains
\(+2.93\%\) over round-robin, while two-stage EV and Ridge-\(M\) trail the
shortcut by 0.81\% and 0.61\% (\Cref{tab:llama-controller-transfer}).  The replication thus preserves both the measured continuation effects
and the measurement-to-control gap.

\begin{table}[!htbp]
\centering
\footnotesize
\setlength{\tabcolsep}{2.8pt}
\begin{tabular}{lccc}
\toprule
Event & Draw & Any of four & Persistence AUROC \\
\midrule
Recovery & 10.2\% & 14.5\% & \shortstack{0.729\\\([0.695,0.763]\)} \\
Damage & 0.9\% & 2.0\% & \shortstack{0.730\\\([0.687,0.768]\)} \\
\bottomrule
\end{tabular}
\caption{Full-protocol Llama-8B replication.  ``Any of four'' denotes
opportunity detection.}
\label{tab:llama-estimands}
\end{table}

\begin{table}[t]
\centering
\footnotesize
\setlength{\tabcolsep}{2.2pt}
\begin{tabular}{lrr}
\toprule
Policy & Gain vs.\ RR & \(\Delta\) vs.\ shortcut \\
\midrule
Round-robin & \shortstack{+0.00\%\\\([+0.00,+0.00]\)} &
\shortstack{-2.85\%\\\([-3.65,-2.07]\)} \\
\rowcolor{resultgreenbg}
Correctness shortcut & \shortstack{+2.93\%\\\([+2.11,+3.79]\)} &
\shortstack{+0.00\%\\\([+0.00,+0.00]\)} \\
Two-stage EV & \shortstack{+2.10\%\\\([+1.54,+2.68]\)} &
\keyresult{\shortstack{-0.81\%\\\([-1.31,-0.31]\)}} \\
Direct Ridge-$M$ & \shortstack{+2.30\%\\\([+1.51,+3.11]\)} &
\keyresult{\shortstack{-0.61\%\\\([-0.91,-0.35]\)}} \\
\bottomrule
\end{tabular}
\caption{Zero-refit transfer to Llama-8B using controllers fitted on the two
primary models.  Brackets are paired 95\% full-system bootstrap intervals.}
\label{tab:llama-controller-transfer}
\end{table}

\section{Discussion}
\label{sec:discussion}

A signal can predict current answer quality, predict continuation effects,
or improve a sequence of compute-allocation decisions.  Our results show
why each claim needs its own evaluation.

\paragraph{Test continuation value beyond error recognition.}
The correctness shortcut captures much of the observable allocation gain,
making it an informative baseline alongside round-robin.  Persistence
retains conditional information about recovery and damage, and highly
persistent correct answers are less vulnerable to harmful revision.  These
findings support using stability as a predictive feature and evaluating its
allocation value through gains over the correctness shortcut.

\paragraph{Charge signals and value later states.}
The fixed-order agreement comparison shows that observation costs can erase
a signal's apparent allocation benefit.  The hindsight comparison on matched
subsets shows that greedy scheduling can still underperform multi-step
optimization even with exact knowledge of immediate value.  The same full
ledger provides a basis for testing stronger sequential controllers using
inference-time features.

\paragraph{Model and dataset differences.}
Ridge-\(M\) gains \(+0.70\%\) over the shortcut on DeepSeek; the Qwen3
increment is inconclusive, and transfer to Llama without refitting gives a
negative gain.  Exploratory replay by dataset shows a DeepSeek gain on
MATH-500 and a loss on GSM8K (\Cref{app:cpu-analysis}).  On the paired
200-problem subset, changing cost weights while fixing action order barely
changes the gain difference between models.  The effects of calibrating
predictions separately for each dataset remain uncertain
(\Cref{app:cpu-control-decomposition}).  The Llama result concerns transfer
of predictors fitted on the two primary models.  Explaining these
differences calls for targeted analysis of recoverable error types, reasoning
changes associated with repair, and predictor transfer.  The audit provides
a common protocol for testing these explanations and comparing stronger
controllers.

\FloatBarrier
\section{Conclusion}

\method{} measures the externally graded value of one more reasoning block
from the same prefix, while accounting for its cost.  Across the evaluated
models, continuation can repair incorrect answers and damage correct ones.
The tested lightweight, myopic controllers add little beyond identifying
likely errors.  This measurement-to-control gap motivates policies that use
inference-time signals to value both immediate outcomes and future reasoning
states.  Their benefit can be tested through accuracy--cost gains over the
correctness shortcut.

\section{Limitations}

The study evaluates three open-weight models on mathematics tasks, with
continuations of 128, 256, and 512 tokens.  Generalization to other tasks,
models, and reasoning actions requires further evaluation.  Continuation
value depends on sampling and on how answers are generated, parsed, and
graded.  Exact-match grading may miss equivalent answers.  Bootstrap
intervals quantify variation across problems while keeping the generated
trajectories fixed.

The controller results cover lightweight policies that score only the next
reasoning step.  In the Llama replication, these policies are transferred
without refitting.  Correctness labels from separately sampled branches
support offline diagnostics.  Stored multi-step transitions are available
only along the main trajectories.  Exact dynamic programming uses these
transitions and known future outcomes to provide an upper bound on sampled
groups of four problems.  The same protocol supports evaluating richer
controllers.

\section{Ethical considerations}

The study uses established mathematics benchmarks and open-weight models,
performs no human-subject intervention, and trains lightweight diagnostic
predictors.  Better allocation could reduce the computation needed for a
given answer quality.  The audit itself requires generation of multiple
spines and continuations, answer probes, and auxiliary scoring.  We report
measured GPU time and alternative cost ledgers in
\Cref{app:cost-sensitivity}.  Energy consumption, emissions, and net
environmental savings fall outside these measurements.

Premature stopping can preserve consequential errors in applications such as
medicine and law, while continuing can damage an initially correct answer.
Applying adaptive stopping in these domains requires task-specific
validation of both recovery and damage, together with safeguards
appropriate to the consequences of errors.

\section*{Acknowledgments}

This work was supported by the National Key Research and Development
Program of China (2025YFF0514401) and the National Natural Science
Foundation of China (62231022).

\FloatBarrier
\bibliography{references}

\begin{thebibliography}{42}
\providecommand{\natexlab}[1]{#1}

\bibitem[{Aggarwal et~al.(2023)Aggarwal, Madaan, Yang, and
  Mausam}]{aggarwal2023adaptive}
Pranjal Aggarwal, Aman Madaan, Yiming Yang, and Mausam. 2023.
\newblock \href {https://arxiv.org/abs/2305.11860} {Let's sample step by step:
  Adaptive-consistency for efficient reasoning and coding with {LLMs}}.
\newblock \emph{arXiv preprint arXiv:2305.11860}.

\bibitem[{Banino et~al.(2021)Banino, Balaguer, and
  Blundell}]{banino2021pondernet}
Andrea Banino, Jan Balaguer, and Charles Blundell. 2021.
\newblock \href {https://arxiv.org/abs/2107.05407} {{PonderNet}: Learning to
  ponder}.
\newblock \emph{arXiv preprint arXiv:2107.05407}.

\bibitem[{Brown et~al.(2024)Brown, Juravsky, Ehrlich, Clark, Le, R{\'e}, and
  Mirhoseini}]{brown2024monkeys}
Bradley Brown, Jordan Juravsky, Ryan Ehrlich, Ronald Clark, Quoc~V. Le,
  Christopher R{\'e}, and Azalia Mirhoseini. 2024.
\newblock \href {https://arxiv.org/abs/2407.21787} {Large language monkeys:
  Scaling inference compute with repeated sampling}.
\newblock \emph{arXiv preprint arXiv:2407.21787}.

\bibitem[{Chen et~al.(2025)Chen, Xu, Liang, He, Pang, Yu, Song, Liu, Zhou,
  Zhang, Wang, Tu, Mi, and Yu}]{chen2025overthinking}
Xingyu Chen, Jiahao Xu, Tian Liang, Zhiwei He, Jianhui Pang, Dian Yu, Linfeng
  Song, Qiuzhi Liu, Mengfei Zhou, Zhuosheng Zhang, Rui Wang, Zhaopeng Tu,
  Haitao Mi, and Dong Yu. 2025.
\newblock \href {https://proceedings.mlr.press/v267/chen25bx.html} {Do {NOT}
  think that much for 2+3=? {On} the overthinking of long reasoning models}.
\newblock In \emph{Proceedings of the 42nd International Conference on Machine
  Learning}, volume 267 of \emph{Proceedings of Machine Learning Research},
  pages 9487--9499. PMLR.

\bibitem[{Cobbe et~al.(2021)Cobbe, Kosaraju, Bavarian, Chen, Jun, Kaiser,
  Plappert, Tworek, Hilton, Nakano, Hesse, and Schulman}]{cobbe2021gsm8k}
Karl Cobbe, Vineet Kosaraju, Mohammad Bavarian, Mark Chen, Heewoo Jun, Lukasz
  Kaiser, Matthias Plappert, Jerry Tworek, Jacob Hilton, Reiichiro Nakano,
  Christopher Hesse, and John Schulman. 2021.
\newblock \href {https://arxiv.org/abs/2110.14168} {Training verifiers to solve
  math word problems}.
\newblock \emph{arXiv preprint arXiv:2110.14168}.

\bibitem[{Damani et~al.(2025)Damani, Shenfeld, Peng, Bobu, and
  Andreas}]{damani2025hard}
Mehul Damani, Idan Shenfeld, Andi Peng, Andreea Bobu, and Jacob Andreas. 2025.
\newblock \href {https://openreview.net/pdf?id=6qUUgw9bAZ} {Learning how hard
  to think: Input-adaptive allocation of {LM} computation}.
\newblock In \emph{International Conference on Learning Representations}.

\bibitem[{{DeepSeek-AI}(2025)}]{deepseekai2025r1}
{DeepSeek-AI}. 2025.
\newblock \href {https://arxiv.org/abs/2501.12948} {{DeepSeek-R1}:
  Incentivizing reasoning capability in {LLMs} via reinforcement learning}.
\newblock \emph{arXiv preprint arXiv:2501.12948}.

\bibitem[{Grattafiori et~al.(2024)Grattafiori, Dubey, Jauhri, Pandey, Kadian,
  Al-Dahle, Letman, Mathur, Schelten, Vaughan et~al.}]{grattafiori2024llama3}
Aaron Grattafiori, Abhimanyu Dubey, Abhinav Jauhri, Abhinav Pandey, Abhishek
  Kadian, Ahmad Al-Dahle, Aiesha Letman, Akhil Mathur, Alan Schelten, Alex
  Vaughan, et~al. 2024.
\newblock \href {https://arxiv.org/abs/2407.21783} {The {Llama 3} herd of
  models}.
\newblock \emph{arXiv preprint arXiv:2407.21783}.

\bibitem[{Graves(2016)}]{graves2016act}
Alex Graves. 2016.
\newblock \href {https://arxiv.org/abs/1603.08983} {Adaptive computation time
  for recurrent neural networks}.
\newblock \emph{arXiv preprint arXiv:1603.08983}.

\bibitem[{Hay et~al.(2012)Hay, Russell, Tolpin, and Shimony}]{hay2012selecting}
Nicholas Hay, Stuart Russell, David Tolpin, and Solomon~Eyal Shimony. 2012.
\newblock \href {https://arxiv.org/abs/1207.5879} {Selecting computations:
  Theory and applications}.
\newblock \emph{arXiv preprint arXiv:1207.5879}.

\bibitem[{Hendrycks et~al.(2021)Hendrycks, Burns, Kadavath, Arora, Basart,
  Tang, Song, and Steinhardt}]{hendrycks2021math}
Dan Hendrycks, Collin Burns, Saurav Kadavath, Akul Arora, Steven Basart, Eric
  Tang, Dawn Song, and Jacob Steinhardt. 2021.
\newblock \href
  {https://datasets-benchmarks-proceedings.neurips.cc/paper_files/paper/2021/hash/be83ab3ecd0db773eb2dc1b0a17836a1-Abstract-round2.html}
  {Measuring mathematical problem solving with the {MATH} dataset}.
\newblock In \emph{Proceedings of the Neural Information Processing Systems
  Track on Datasets and Benchmarks}, volume~1.

\bibitem[{Horvitz et~al.(1989)Horvitz, Cooper, and
  Heckerman}]{horvitz1989reflection}
Eric~J. Horvitz, Gregory~F. Cooper, and David~E. Heckerman. 1989.
\newblock \href {https://ijcai.org/proceedings/1989/0180.pdf} {Reflection and
  action under scarce resources: Theoretical principles and empirical study}.
\newblock In \emph{International Joint Conference on Artificial Intelligence},
  pages 1121--1127.

\bibitem[{Huang et~al.(2024)Huang, Chen, Mishra, Zheng, Yu, Song, and
  Zhou}]{huang2024selfcorrect}
Jie Huang, Xinyun Chen, Swaroop Mishra, Huaixiu~Steven Zheng, Adams~Wei Yu,
  Xinying Song, and Denny Zhou. 2024.
\newblock \href {https://openreview.net/forum?id=IkmD3fKBPQ} {Large language
  models cannot self-correct reasoning yet}.
\newblock In \emph{International Conference on Learning Representations}.

\bibitem[{Iwase et~al.(2026)Iwase, Ichihara, Quamar, and
  Komiyama}]{iwase2026prefix}
Naoto Iwase, Yuki Ichihara, Mohammad~Atif Quamar, and Junpei Komiyama. 2026.
\newblock \href {https://arxiv.org/abs/2605.07654} {Reliable chain-of-thought
  via prefix consistency}.
\newblock \emph{arXiv preprint arXiv:2605.07654}.

\bibitem[{Jindal et~al.(2026)Jindal, Akuthota, Taneja, and
  Sharma}]{jindal2026polr}
Ishan Jindal, Sai~Prashanth Akuthota, Jayant Taneja, and Sachin~Dev Sharma.
  2026.
\newblock \href {https://arxiv.org/abs/2601.21494} {The path of least
  resistance: Guiding {LLM} reasoning trajectories with prefix consensus}.
\newblock \emph{arXiv preprint arXiv:2601.21494}.

\bibitem[{Kadavath et~al.(2022)Kadavath, Conerly, Askell, Henighan, Drain,
  Perez, Schiefer, Hatfield-Dodds, DasSarma, Tran-Johnson
  et~al.}]{kadavath2022know}
Saurav Kadavath, Tom Conerly, Amanda Askell, Tom Henighan, Dawn Drain, Ethan
  Perez, Nicholas Schiefer, Zac Hatfield-Dodds, Nova DasSarma, Eli
  Tran-Johnson, et~al. 2022.
\newblock \href {https://arxiv.org/abs/2207.05221} {Language models (mostly)
  know what they know}.
\newblock \emph{arXiv preprint arXiv:2207.05221}.

\bibitem[{Kojima et~al.(2022)Kojima, Gu, Reid, Matsuo, and
  Iwasawa}]{kojima2022zeroshot}
Takeshi Kojima, Shixiang~Shane Gu, Machel Reid, Yutaka Matsuo, and Yusuke
  Iwasawa. 2022.
\newblock \href
  {https://proceedings.neurips.cc/paper_files/paper/2022/hash/8bb0d291acd4acf06ef112099c16f326-Abstract-Conference.html}
  {Large language models are zero-shot reasoners}.
\newblock In \emph{Advances in Neural Information Processing Systems}.

\bibitem[{Kuhn et~al.(2023)Kuhn, Gal, and
  Farquhar}]{kuhn2023semanticuncertainty}
Lorenz Kuhn, Yarin Gal, and Sebastian Farquhar. 2023.
\newblock \href {https://arxiv.org/abs/2302.09664} {Semantic uncertainty:
  Linguistic invariances for uncertainty estimation in natural language
  generation}.
\newblock \emph{arXiv preprint arXiv:2302.09664}.

\bibitem[{Kumar et~al.(2025)Kumar, Zhuang, Agarwal, Su, Co-Reyes, Singh,
  Baumli, Iqbal, Bishop, Roelofs, Zhang, McKinney, Shrivastava, Paduraru,
  Tucker, Precup, Behbahani, and Faust}]{kumar2025score}
Aviral Kumar, Vincent Zhuang, Rishabh Agarwal, Yi~Su, JD~Co-Reyes, Avi Singh,
  Kate Baumli, Shariq Iqbal, Colton Bishop, Rebecca Roelofs, Lei~M. Zhang, Kay
  McKinney, Disha Shrivastava, Cosmin Paduraru, George Tucker, Doina Precup,
  Feryal Behbahani, and Aleksandra Faust. 2025.
\newblock \href
  {https://proceedings.iclr.cc/paper_files/paper/2025/hash/871ac99fdc5282d0301934d23945ebaa-Abstract-Conference.html}
  {Training language models to self-correct via reinforcement learning}.
\newblock In \emph{International Conference on Learning Representations}.

\bibitem[{Li et~al.(2026)Li, He, Li, Chai, Yu, and Yuan}]{li2026trace}
Jiakun Li, Xingwei He, Kefan Li, Hongzheng Chai, Hongyue Yu, and Yuan Yuan.
  2026.
\newblock \href {https://arxiv.org/abs/2604.17304} {Efficient test-time scaling
  via temporal reasoning aggregation}.
\newblock \emph{arXiv preprint arXiv:2604.17304}.

\bibitem[{Lightman et~al.(2024)Lightman, Kosaraju, Burda, Edwards, Baker, Lee,
  Leike, Schulman, Sutskever, and Cobbe}]{lightman2023verify}
Hunter Lightman, Vineet Kosaraju, Yura Burda, Harri Edwards, Bowen Baker, Teddy
  Lee, Jan Leike, John Schulman, Ilya Sutskever, and Karl Cobbe. 2024.
\newblock \href {https://openreview.net/pdf?id=v8L0pN6EOi} {Let's verify step
  by step}.
\newblock In \emph{International Conference on Learning Representations}.

\bibitem[{Madaan et~al.(2023)Madaan, Tandon, Gupta, Hallinan, Gao, Wiegreffe,
  Alon, Dziri, Prabhumoye, Yang, Gupta, Majumder, Hermann, Welleck,
  Yazdanbakhsh, and Clark}]{madaan2023selfrefine}
Aman Madaan, Niket Tandon, Prakhar Gupta, Skyler Hallinan, Luyu Gao, Sarah
  Wiegreffe, Uri Alon, Nouha Dziri, Shrimai Prabhumoye, Yiming Yang, Shashank
  Gupta, Bodhisattwa~Prasad Majumder, Katherine Hermann, Sean Welleck, Amir
  Yazdanbakhsh, and Peter Clark. 2023.
\newblock \href
  {https://proceedings.neurips.cc/paper_files/paper/2023/hash/91edff07232fb1b55a505a9e9f6c0ff3-Abstract-Conference.html}
  {{Self-Refine}: Iterative refinement with self-feedback}.
\newblock In \emph{Advances in Neural Information Processing Systems}.

\bibitem[{Min et~al.(2026)Min, Vaccarino, Chen, Wu, Yona, and
  Cheng}]{min2026puma}
Dehai Min, Giovanni Vaccarino, Huiyi Chen, Yongliang Wu, Gal Yona, and
  Lu~Cheng. 2026.
\newblock \href {https://arxiv.org/abs/2605.17672} {Stop when reasoning
  converges: Semantic-preserving early exit for reasoning models}.
\newblock \emph{arXiv preprint arXiv:2605.17672}.

\bibitem[{Muennighoff et~al.(2025)Muennighoff, Yang, Shi, Li, Fei-Fei,
  Hajishirzi, Zettlemoyer, Liang, Cand{\`e}s, and
  Hashimoto}]{muennighoff2025s1}
Niklas Muennighoff, Zitong Yang, Weijia Shi, Xiang~Lisa Li, Li~Fei-Fei,
  Hannaneh Hajishirzi, Luke Zettlemoyer, Percy Liang, Emmanuel Cand{\`e}s, and
  Tatsunori Hashimoto. 2025.
\newblock \href {https://arxiv.org/abs/2501.19393} {{s1}: Simple test-time
  scaling}.
\newblock \emph{arXiv preprint arXiv:2501.19393}.

\bibitem[{Nguyen et~al.(2026)Nguyen, Gupta, and Le}]{nguyen2026uab}
Manh Nguyen, Sunil Gupta, and Hung Le. 2026.
\newblock \href {https://arxiv.org/abs/2605.26849v2} {Uncertainty-aware budget
  allocation for adaptive test-time reasoning}.
\newblock \emph{arXiv preprint arXiv:2605.26849}.

\bibitem[{Pan et~al.(2025)Pan, Li, Lian, Snell, Zhou, Yala, Darrell, Keutzer,
  and Suhr}]{pan2025apr}
Jiayi Pan, Xiuyu Li, Long Lian, Charlie Snell, Yifei Zhou, Adam Yala, Trevor
  Darrell, Kurt Keutzer, and Alane Suhr. 2025.
\newblock \href {https://openreview.net/attachment?id=YgwQ7sXPXU&name=pdf}
  {Learning adaptive parallel reasoning with language models}.
\newblock In \emph{Conference on Language Modeling}.

\bibitem[{Scalena et~al.(2025)Scalena, Zotos, Fersini, Nissim, and
  {\"U}st{\"u}n}]{scalena2025eager}
Daniel Scalena, Leonidas Zotos, Elisabetta Fersini, Malvina Nissim, and Ahmet
  {\"U}st{\"u}n. 2025.
\newblock \href {https://arxiv.org/abs/2510.11170v2} {{EAGer}: Entropy-aware
  {GEneRation} for adaptive inference-time scaling}.
\newblock \emph{arXiv preprint arXiv:2510.11170}.

\bibitem[{Schuster et~al.(2022)Schuster, Fisch, Gupta, Dehghani, Bahri, Tran,
  Tay, and Metzler}]{schuster2022calm}
Tal Schuster, Adam Fisch, Jai Gupta, Mostafa Dehghani, Dara Bahri, Vinh~Q.
  Tran, Yi~Tay, and Donald Metzler. 2022.
\newblock \href
  {https://proceedings.neurips.cc/paper_files/paper/2022/hash/6fac9e316a4ae75ea244ddcef1982c71-Abstract-Conference.html}
  {Confident adaptive language modeling}.
\newblock In \emph{Advances in Neural Information Processing Systems}.

\bibitem[{Setlur et~al.(2024)Setlur, Nagpal, Fisch, Geng, Eisenstein, Agarwal,
  Agarwal, Berant, and Kumar}]{setlur2024progress}
Amrith Setlur, Chirag Nagpal, Adam Fisch, Xinyang Geng, Jacob Eisenstein,
  Rishabh Agarwal, Alekh Agarwal, Jonathan Berant, and Aviral Kumar. 2024.
\newblock \href {https://arxiv.org/abs/2410.08146} {Rewarding progress: Scaling
  automated process verifiers for {LLM} reasoning}.
\newblock \emph{arXiv preprint arXiv:2410.08146}.

\bibitem[{Smirnov et~al.(2026)Smirnov, Nguyen, Senichev, Ta
  et~al.}]{smirnov2026thinkbooster}
Vladislav Smirnov, Chieu Nguyen, Sergey Senichev, Minh~Ngoc Ta, et~al. 2026.
\newblock \href {https://arxiv.org/abs/2606.06915} {{ThinkBooster}: A unified
  framework for seamless test-time scaling of {LLM} reasoning}.
\newblock \emph{arXiv preprint arXiv:2606.06915}.

\bibitem[{Snell et~al.(2024)Snell, Lee, Xu, and Kumar}]{snell2024scaling}
Charlie Snell, Jaehoon Lee, Kelvin Xu, and Aviral Kumar. 2024.
\newblock \href {https://arxiv.org/abs/2408.03314} {Scaling {LLM} test-time
  compute optimally can be more effective than scaling model parameters}.
\newblock \emph{arXiv preprint arXiv:2408.03314}.

\bibitem[{Uesato et~al.(2022)Uesato, Kushman, Kumar, Song, Siegel, Wang,
  Creswell, Irving, and Higgins}]{uesato2022processoutcome}
Jonathan Uesato, Nate Kushman, Ramana Kumar, Francis Song, Noah Siegel, Lisa
  Wang, Antonia Creswell, Geoffrey Irving, and Irina Higgins. 2022.
\newblock \href {https://arxiv.org/abs/2211.14275} {Solving math word problems
  with process- and outcome-based feedback}.
\newblock \emph{arXiv preprint arXiv:2211.14275}.

\bibitem[{Wang et~al.(2023{\natexlab{a}})Wang, Li, Shao, Xu, Dai, Li, Chen, Wu,
  and Sui}]{wang2023mathshepherd}
Peiyi Wang, Lei Li, Zhihong Shao, R.~X. Xu, Damai Dai, Yifei Li, Deli Chen,
  Y.~Wu, and Zhifang Sui. 2023{\natexlab{a}}.
\newblock \href {https://arxiv.org/abs/2312.08935} {{Math-Shepherd}: Verify and
  reinforce {LLMs} step-by-step without human annotations}.
\newblock \emph{arXiv preprint arXiv:2312.08935}.

\bibitem[{Wang et~al.(2023{\natexlab{b}})Wang, Wei, Schuurmans, Le, Chi,
  Narang, Chowdhery, and Zhou}]{wang2023selfconsistency}
Xuezhi Wang, Jason Wei, Dale Schuurmans, Quoc Le, Ed~H. Chi, Sharan Narang,
  Aakanksha Chowdhery, and Denny Zhou. 2023{\natexlab{b}}.
\newblock \href {https://arxiv.org/abs/2203.11171v4} {Self-consistency improves
  chain of thought reasoning in language models}.
\newblock In \emph{International Conference on Learning Representations}.

\bibitem[{Wei et~al.(2022)Wei, Wang, Schuurmans, Bosma, Ichter, Xia, Chi, Le,
  and Zhou}]{wei2022chain}
Jason Wei, Xuezhi Wang, Dale Schuurmans, Maarten Bosma, Brian Ichter, Fei Xia,
  Ed~H. Chi, Quoc~V. Le, and Denny Zhou. 2022.
\newblock \href
  {https://proceedings.neurips.cc/paper/2022/hash/9d5609613524ecf4f15af0f7b31abca4-Abstract-Conference.html}
  {Chain-of-thought prompting elicits reasoning in large language models}.
\newblock In \emph{Advances in Neural Information Processing Systems}.

\bibitem[{Wu et~al.(2025)Wu, Sun, Li, Welleck, and
  Yang}]{wu2025inferencescaling}
Yangzhen Wu, Zhiqing Sun, Shanda Li, Sean Welleck, and Yiming Yang. 2025.
\newblock \href
  {https://proceedings.iclr.cc/paper_files/paper/2025/hash/8c3caae2f725c8e2a55ecd600563d172-Abstract-Conference.html}
  {Inference scaling laws: An empirical analysis of compute-optimal inference
  for {LLM} problem-solving}.
\newblock In \emph{International Conference on Learning Representations}.

\bibitem[{Xiong et~al.(2024)Xiong, Hu, Lu, Li, Fu, He, and
  Hooi}]{xiong2024uncertainty}
Miao Xiong, Zhiyuan Hu, Xinyang Lu, Yifei Li, Jie Fu, Junxian He, and Bryan
  Hooi. 2024.
\newblock \href {https://openreview.net/pdf?id=gjeQKFxFpZ} {Can {LLMs} express
  their uncertainty? an empirical evaluation of confidence elicitation in
  {LLMs}}.
\newblock In \emph{International Conference on Learning Representations}.

\bibitem[{Yang et~al.(2025{\natexlab{a}})Yang, Li, Yang, Zhang, Hui, Zheng, Yu,
  Gao, Huang, Lv et~al.}]{yang2025qwen3}
An~Yang, Anfeng Li, Baosong Yang, Beichen Zhang, Binyuan Hui, Bo~Zheng, Bowen
  Yu, Chang Gao, Chengen Huang, Chenxu Lv, et~al. 2025{\natexlab{a}}.
\newblock \href {https://arxiv.org/abs/2505.09388} {{Qwen3} technical report}.
\newblock \emph{arXiv preprint arXiv:2505.09388}.

\bibitem[{Yang et~al.(2025{\natexlab{b}})Yang, Si, Duan, Zhu, Zhu, Li, Chen,
  Lin, and Wang}]{yang2025dynamicexit}
Chenxu Yang, Qingyi Si, Yongjie Duan, Zheliang Zhu, Chenyu Zhu, Qiaowei Li,
  Minghui Chen, Zheng Lin, and Weiping Wang. 2025{\natexlab{b}}.
\newblock \href {https://arxiv.org/abs/2504.15895v3} {Dynamic early exit in
  reasoning models}.
\newblock \emph{arXiv preprint arXiv:2504.15895}.

\bibitem[{Yao et~al.(2023)Yao, Yu, Zhao, Shafran, Griffiths, Cao, and
  Narasimhan}]{yao2023tot}
Shunyu Yao, Dian Yu, Jeffrey Zhao, Izhak Shafran, Thomas~L. Griffiths, Yuan
  Cao, and Karthik Narasimhan. 2023.
\newblock \href
  {https://proceedings.neurips.cc/paper/2023/hash/271db9922b8d1f4dd7aaef84ed5ac703-Abstract.html}
  {Tree of thoughts: Deliberate problem solving with large language models}.
\newblock In \emph{Advances in Neural Information Processing Systems}.

\bibitem[{Zhang et~al.(2025)Zhang, Zheng, Wu, Zhang, Lin, Yu, Liu, Zhou, and
  Lin}]{zhang2025prm}
Zhenru Zhang, Chujie Zheng, Yangzhen Wu, Beichen Zhang, Runji Lin, Bowen Yu,
  Dayiheng Liu, Jingren Zhou, and Junyang Lin. 2025.
\newblock \href {https://arxiv.org/abs/2501.07301} {The lessons of developing
  process reward models in mathematical reasoning}.
\newblock \emph{arXiv preprint arXiv:2501.07301}.

\bibitem[{Zilberstein(1996)}]{zilberstein1996anytime}
Shlomo Zilberstein. 1996.
\newblock \href
  {https://onlinelibrary.wiley.com/doi/abs/10.1609/aimag.v17i3.1232} {Using
  anytime algorithms in intelligent systems}.
\newblock \emph{AI Magazine}, 17(3):73--83.

\end{thebibliography}
\clearpage
\appendix
\noindent\textbf{Appendix roadmap.}\quad
\Cref{app:experimental-details} gives the complete intervention, labels,
proxies, and statistical protocol.
\Cref{app:replay-controls} details replay units, observability, leakage
controls, and non-myopic optimization.
\Cref{app:proxy-detail} reports the complete proxy and controller diagnostics.
\Cref{app:intervention-robustness} contains depth, branch-count, length,
answer-cue, and parser analyses.
\Cref{app:external-replication} gives the Llama replication and physical-cost
sensitivity.
\Cref{app:cpu-analysis} reports exploratory analyses of dataset composition,
score calibration, cost weighting, recovery, and automatic readout.

\section{Experimental Details}
\label{app:experimental-details}

\subsection{Fixed protocol}

The primary campaign evaluates Qwen3-8B and
DeepSeek-R1-Distill-Qwen-7B on the 1,319-problem GSM8K test set and all
500 MATH-500 problems.  Reasoning uses temperature \(0.6\) and
top-\(p=0.95\).  A deterministic problem-ID partition assigns 80\% of
problems to analysis and 20\% to confirmation; no problem crosses the split.
Unless specified otherwise, uncertainty intervals use 2,000 problem-level
bootstrap replicates.  The off-spine correspondence control uses 1,000
within-stratum permutations.

For each model--problem pair, four independent spines are sampled to a
4,096-token cap.  Prefixes are frozen at depths
\(256,512,\ldots,4096\).  Each nonterminal prefix receives an answer-now probe,
one continuation reusing the next stored on-spine segment, and three
independently seeded off-spine continuations.  Every branch contains exactly
256 reasoning tokens unless the model has already emitted its terminal answer,
and every endpoint receives the same greedy suffix
\texttt{Therefore, the final answer is} \verb|\boxed{|.  Decoding stops at the
closing brace or after 32 answer tokens.  The on-spine branch must reproduce
the stored successor token segment exactly; off-spine branches never re-enter
the spine.

This design yields 232,832 stored checkpoints, 218,280 eligible nonterminal
checkpoints, and 873,120 branch references.  Reusing the on-spine segment
avoids redundant physical generation while retaining its full reasoning and
answer-probe cost in \(L_{s,1}\).

\subsection{Answer extraction and target construction}

GSM8K evaluation extracts the final numeric answer and canonicalizes signs,
commas, decimals, and fractions.  MATH-500 evaluation extracts the final boxed
expression, applies deterministic symbolic normalization, and falls back to
normalized string equality.  Unparseable answers remain in the denominator
and are incorrect.  Raw output, extracted span, canonical form, parser status,
and exact-match bit are retained for every endpoint.

A checkpoint enters Eq.~\ref{eq:value} only when all four continuation labels
and positive logical costs exist.  We define
\[
\begin{aligned}
  y_s^+&=\mathbf 1[\ms>0] && (S_s=0),\\
  y_s^-&=\mathbf 1[\ms<0] && (S_s=1).
\end{aligned}
\]
The numerator of Eq.~\ref{eq:value} is mean branch correctness minus stop
correctness, and the denominator is mean branch cost.  The factor 1,000
expresses expected correctness change per 1,000 logical tokens.  The ratio
of means differs from the mean of branch-specific ratios;
\Cref{tab:value-ratio-sensitivity} shows that the two are empirically nearly
identical here because reasoning length is fixed and only the short answer
length varies.

\subsection{Observable proxies}

\begin{description}[style=nextline,itemsep=2pt]
  \item[Persistence.] Fraction of up to four most recent stop answers equal to
  the current canonical answer.  It is structurally missing at the first
  checkpoint.
  \item[Answer confidence and ANLL.] Exponential mean selected-token log
  probability over the answer span, and its negative-log counterpart.
  \item[Modal agreement.] Fraction of four equal-depth spines supporting the
  lexicographically tie-broken modal answer.  It is an offline cross-seed
  diagnostic unless all four trajectories are explicitly charged.
  \item[Process reward.] Minimum positive-class probability across
  double-newline-delimited steps from Qwen2.5-Math-PRM-7B
  ~\citep{zhang2025prm}.
  \item[Entropy.] Mean entropy of the top-20 answer-token probabilities plus a
  residual-mass bucket.
\end{description}

Missing proxy values are not imputed from correctness or continuation labels.
Available-case denominators are reported for every proxy.  PRM scores can be
missing when no valid step separator remains; confidence and entropy can be
missing when log probabilities are unavailable; agreement can be missing when
no spine answer parses.

\subsection{Statistical protocol}

Analysis problems choose proxy orientation, tie-preserving bins,
Jeffreys-smoothed calibration probabilities, persistence thresholds, feature
scaling, imputation, regularization, and controller coefficients.
Confirmation problems provide every reported held-out target and replay
result.  No problem crosses the split.

Recovery metrics condition on \(S_s=0\); damage metrics condition on \(S_s=1\).
Average precision updates precision and recall only after adding a complete
equal-score block, preventing record order from breaking ties.  AP is reported
with confirmation prevalence and AP lift.  Brier skill compares with an
analysis-prevalence constant predictor:
\[
 \mathrm{BSS}=1-\frac{\mathrm{BS}}
 {\frac1n\sum_i(y_i-\pi_{\mathrm{analysis}})^2}.
\]
Problem-bootstrap intervals resample problem IDs together with all nested
models, spines, checkpoints, and branches.  They quantify between-problem
uncertainty conditional on the stored trajectories.  Replay intervals use
full-system resampling or fixed-order per-problem area contributions, as
specified in each table (\Cref{app:replay-controls}).  The DP analysis
resamples four-problem subsets, as described with its results.

Difficulty is \(1-\)mean \(t=256\) stop accuracy pooled over the two primary
models and four spines.  It is used only for diagnostic stratification, never
as a controller feature.  Proxy bins are fitted on analysis problems without
splitting ties.  Collision cells hold model, dataset, depth, difficulty,
current correctness, proxy, and proxy bin fixed, and are retained only when
both the feasible nonzero event and a zero-value control occur.  Because this
coverage depends on bin resolution, we also report exact matching for discrete
proxies, 5/10/20/50-bin sensitivity, conditional AUROC, nearest-neighbor
disagreement, and out-of-sample covariate-only log-loss reduction.  Difficulty
deletion reruns all grouping and fitted diagnostic baselines without that
label-aware stratifier.

\subsection{Cost coordinates}
\label{app:cost-coordinates}

The logical ledger charges prompt tokens, generated reasoning, answer probes,
and any proxy-specific scoring needed before a policy can act.  Prompt tokens
are deduplicated when a shared KV state can be reused; all generated reasoning
and answer tokens are charged.  A physical sidecar records role-specific
input/output tokens, GPU seconds, peak memory, and throughput.  Estimated FLOPs
account for causal sequence length, grouped-query attention, SwiGLU
projections, vocabulary heads, and reusable prefix KV states.  Calibrated H20
seconds apply separate observed rates to reasoning generation, answer probes,
and PRM prefill.  We report these resource measures separately.

\FloatBarrier
\section{Replay and Identification Controls}
\label{app:replay-controls}

\subsection{Problem-level replay}

The primary replay runs four campaigns, one per spine seed.  Each campaign has
one frontier for each of 372 confirmation problems, so its accuracy is
problem-level exact match.  At each ledger boundary, a policy selects one
frontier, pays the stored transition cost, and reveals its successor answer.
The common endpoint for a policy comparison is the largest budget reachable
by every policy.  For repetition \(r\), normalized AULC is
\[
 \frac{1}{B_r-B_{0,r}}\int_{B_{0,r}}^{B_r}\mathrm{Acc}_r(b)\,db,
\]
using the right-continuous accuracy curve.  The main controller hierarchy uses
full-system bootstrap: each replicate reconstructs the resampled problem pool,
creates independent copies for duplicate draws, reruns all four global
orderings, and recomputes the common endpoint.

\begin{table*}[t]
\centering
\footnotesize
\resizebox{\linewidth}{!}{%
\begin{tabular}{llrrr}
\toprule
Model & Policy & Mean normalized AULC & Rel.\ normalized-AULC gain vs.\ RR (95\% CI) & Final problem accuracy \\
\midrule
Qwen3 & Fixed 1,024 & 0.820 & -3.66\% [-4.34, -3.01] & 0.909 \\
Qwen3 & Round-robin & 0.851 & +0.00\% [+0.00, +0.00] & 0.909 \\
Qwen3 & TRACE-style & 0.814 & -4.29\% [-5.96, -2.54] & 0.909 \\
Qwen3 & PRM & 0.439 & -48.41\% [-52.94, -43.76] & 0.476 \\
Qwen3 & ANLL & 0.834 & -2.00\% [-3.52, -0.44] & 0.909 \\
Qwen3 & Entropy & 0.832 & -2.27\% [-3.78, -0.70] & 0.909 \\
\addlinespace
DeepSeek & Fixed 1,024 & 0.792 & -2.58\% [-3.21, -2.00] & 0.847 \\
DeepSeek & Round-robin & 0.813 & +0.00\% [+0.00, +0.00] & 0.847 \\
DeepSeek & TRACE-style & 0.813 & +0.04\% [-1.06, +1.21] & 0.847 \\
DeepSeek & PRM & 0.537 & -33.88\% [-37.99, -29.68] & 0.554 \\
DeepSeek & ANLL & 0.823 & +1.33\% [+0.35, +2.38] & 0.847 \\
DeepSeek & Entropy & 0.823 & +1.30\% [+0.27, +2.44] & 0.847 \\
\addlinespace
\bottomrule
\end{tabular}
}
\caption{Problem-level equal-ledger replay over 372 confirmation problems,
with one spine per problem in each of four repetitions.  Mean normalized AULC
averages the normalized areas over repetitions.  Paired intervals use 2,000
problem-cluster bootstrap draws of per-problem area contributions, retaining
all four seeds and the fixed full-universe replay order.  Label-aware
opportunity diagnostics appear in Table~\ref{tab:opportunity-replay}; charged
cross-spine agreement appears in Table~\ref{tab:agreement-accounting}.}
\label{tab:spine-replicate-replay}
\end{table*}

\Cref{tab:spine-replicate-replay} reports the resulting problem-level replay
for the predefined non-label-aware baselines.

\subsection{Successor leakage and independent branches}

Branch 1 is the next stored spine segment, so its correctness label equals
the successor checkpoint's answer label.  The mixed four-branch score includes
this realized outcome.  The off-spine score uses branches 2--4 to rank the
branch-1 transition without observing its realized outcome.  This offline diagnostic tests
whether independent continuations predict the successor; its continuation
correctness labels require grading and are unavailable to an online policy.

\begin{table*}[t]
\centering
\footnotesize
\begin{tabular}{llrrr}
\toprule
Model & Label-aware ordering & Mean normalized AULC & Gain vs.\ RR (95\% CI) & Final problem accuracy \\
\midrule
Qwen3 & Round-robin & 0.851 & +0.00\% [+0.00, +0.00] & 0.909 \\
Qwen3 & On-spine hindsight & 0.842 & -1.10\% [-3.32, +1.33] & 0.909 \\
Qwen3 & Mixed myopic & 0.869 & +2.07\% [+0.19, +4.02] & 0.909 \\
\rowcolor{resultgreenbg}
Qwen3 & Off-spine myopic & 0.863 & +1.44\% [-0.48, +3.37] & 0.909 \\
\addlinespace
DeepSeek & Round-robin & 0.813 & +0.00\% [+0.00, +0.00] & 0.847 \\
DeepSeek & On-spine hindsight & 0.844 & +3.88\% [+1.96, +5.90] & 0.847 \\
DeepSeek & Mixed myopic & 0.864 & +6.35\% [+4.52, +8.38] & 0.847 \\
\rowcolor{resultgreenbg}
DeepSeek & Off-spine myopic & 0.856 & +5.41\% [+3.61, +7.29] & 0.847 \\
\addlinespace
\bottomrule
\end{tabular}
\caption{Offline opportunity diagnostics under the single-spine logical
ledger.  On-spine ordering uses the realized successor label; mixed ordering
includes it among four labels; off-spine ordering uses three independent
continuation labels.  Green highlights the off-spine diagnostic.  These scores
require future correctness labels for offline evaluation.  Paired intervals
use 2,000 problem-cluster bootstrap draws, retaining all four spine
repetitions and conditioning on the fixed full-universe replay order.}
\label{tab:opportunity-replay}
\end{table*}

\Cref{tab:opportunity-replay} contrasts that independent ordering with the
mixed and on-spine hindsight orderings under the same ledger.

Leave-one-branch-out evaluation predicts each branch from the other three.
Holding out branch 1 is the direct test of whether independent off-spine
outcomes predict the actual replay successor; holding out branches 2--4 checks
that predictability is not peculiar to the distinguished on-spine branch.

\begin{table*}[t]
\centering
\scriptsize
\setlength{\tabcolsep}{3.2pt}
\begin{tabular}{lllrrrrr}
\toprule
Model & Target & Held-out branch & $n$ & Event rate & AUROC (95\% CI) & AUPRC (95\% CI) & Brier (95\% CI) \\
\midrule
Qwen3 & Beneficial & Pooled 1--4 & 14,744 & 26.6\% & 0.943 [0.929, 0.954] & 0.865 [0.838, 0.890] & 0.061 [0.049, 0.074] \\
Qwen3 & Beneficial & 1 (on-spine) & 3,686 & 26.9\% & 0.941 [0.926, 0.954] & 0.867 [0.838, 0.894] & 0.062 [0.049, 0.075] \\
Qwen3 & Beneficial & 2 (off-spine) & 3,686 & 26.2\% & 0.945 [0.930, 0.958] & 0.863 [0.834, 0.891] & 0.061 [0.049, 0.075] \\
Qwen3 & Beneficial & 3 (off-spine) & 3,686 & 26.0\% & 0.945 [0.930, 0.959] & 0.859 [0.826, 0.889] & 0.061 [0.049, 0.076] \\
Qwen3 & Beneficial & 4 (off-spine) & 3,686 & 27.2\% & 0.940 [0.923, 0.954] & 0.872 [0.841, 0.900] & 0.061 [0.049, 0.076] \\
\addlinespace
Qwen3 & Harmful & Pooled 1--4 & 74,536 & 1.4\% & 0.849 [0.810, 0.880] & 0.490 [0.381, 0.580] & 0.009 [0.007, 0.011] \\
Qwen3 & Harmful & 1 (on-spine) & 18,634 & 1.4\% & 0.859 [0.822, 0.889] & 0.486 [0.381, 0.579] & 0.009 [0.007, 0.011] \\
Qwen3 & Harmful & 2 (off-spine) & 18,634 & 1.3\% & 0.848 [0.800, 0.885] & 0.478 [0.359, 0.580] & 0.009 [0.007, 0.011] \\
Qwen3 & Harmful & 3 (off-spine) & 18,634 & 1.4\% & 0.845 [0.796, 0.882] & 0.473 [0.353, 0.579] & 0.009 [0.007, 0.011] \\
Qwen3 & Harmful & 4 (off-spine) & 18,634 & 1.5\% & 0.846 [0.804, 0.883] & 0.526 [0.415, 0.624] & 0.009 [0.007, 0.011] \\
\addlinespace
DeepSeek & Beneficial & Pooled 1--4 & 17,604 & 18.9\% & 0.944 [0.934, 0.954] & 0.839 [0.812, 0.864] & 0.048 [0.040, 0.057] \\
DeepSeek & Beneficial & 1 (on-spine) & 4,401 & 18.8\% & 0.948 [0.935, 0.960] & 0.857 [0.826, 0.886] & 0.045 [0.037, 0.054] \\
DeepSeek & Beneficial & 2 (off-spine) & 4,401 & 18.8\% & 0.941 [0.928, 0.954] & 0.824 [0.790, 0.857] & 0.051 [0.042, 0.061] \\
DeepSeek & Beneficial & 3 (off-spine) & 4,401 & 18.9\% & 0.947 [0.935, 0.960] & 0.841 [0.810, 0.872] & 0.048 [0.039, 0.057] \\
DeepSeek & Beneficial & 4 (off-spine) & 4,401 & 19.0\% & 0.942 [0.929, 0.954] & 0.833 [0.800, 0.864] & 0.050 [0.041, 0.060] \\
\addlinespace
DeepSeek & Harmful & Pooled 1--4 & 71,676 & 1.9\% & 0.879 [0.858, 0.898] & 0.579 [0.510, 0.643] & 0.010 [0.008, 0.012] \\
DeepSeek & Harmful & 1 (on-spine) & 17,919 & 1.9\% & 0.880 [0.851, 0.904] & 0.586 [0.503, 0.656] & 0.010 [0.008, 0.012] \\
DeepSeek & Harmful & 2 (off-spine) & 17,919 & 1.9\% & 0.884 [0.857, 0.907] & 0.605 [0.535, 0.669] & 0.010 [0.008, 0.011] \\
DeepSeek & Harmful & 3 (off-spine) & 17,919 & 1.9\% & 0.880 [0.853, 0.903] & 0.593 [0.513, 0.665] & 0.010 [0.008, 0.012] \\
DeepSeek & Harmful & 4 (off-spine) & 17,919 & 1.8\% & 0.874 [0.843, 0.903] & 0.534 [0.450, 0.616] & 0.011 [0.009, 0.013] \\
\addlinespace
\bottomrule
\end{tabular}
\caption{Leave-one-branch-out prediction on confirmation checkpoints. The mean event frequency of the other three branches predicts the held-out branch event. Beneficial is conditioned on $S_s=0$ and harmful on $S_s=1$. Branch 1 is the realized replay successor. Intervals resample complete problems 2,000 times.}
\label{tab:leave-one-branch-out}
\end{table*}

\Cref{tab:leave-one-branch-out} reports the pooled predictor and each of the
four leave-one-branch-out evaluations.

A complementary pooled-spine diagnostic places all four spines of each
problem in the same budget pool.  It measures accuracy over problem--spine
pairs; primary replay measures accuracy over problems.  We report the two
analyses separately because they use different allocation and evaluation
units.

\begin{table*}[t]
\centering
\footnotesize
\resizebox{\linewidth}{!}{%
\begin{tabular}{llcrrr}
\toprule
Model & Ordering score & Sees successor? & Mean normalized AULC & AULC gain vs.\ RR & Off-spine permutation $p$ ($+D/-D$) \\
\midrule
Qwen3 & Round-robin & no & 0.851 & +0.00\% & -- \\
Qwen3 & Mixed 4-branch & yes & 0.856 & +0.57\% & -- \\
Qwen3 & On-spine one-step hindsight & yes & 0.827 & -2.82\% & -- \\
Qwen3 & Off-spine independent & no & 0.849 & -0.19\% & 0.001/0.001 \\
\addlinespace
DeepSeek & Round-robin & no & 0.813 & +0.00\% & -- \\
DeepSeek & Mixed 4-branch & yes & 0.848 & +4.37\% & -- \\
DeepSeek & On-spine one-step hindsight & yes & 0.825 & +1.57\% & -- \\
DeepSeek & Off-spine independent & no & 0.840 & +3.39\% & 0.001/0.001 \\
\addlinespace
\bottomrule
\end{tabular}
}
\caption{Spine-level myopic score decomposition with all four spines of each
problem in one budget pool.  Mixed ordering uses all four branch labels;
one-step hindsight uses the realized on-spine successor; off-spine ordering
uses branches 2--4 to rank actions evaluated on branch 1.  Each score is used
for greedy next-step allocation.  The off-spine permutation test measures
checkpoint correspondence within matching strata; $+D$ includes pooled
difficulty and $-D$ removes it.}
\label{tab:oracle-decomposition}
\end{table*}

\begin{figure*}[t]
\centering
\includegraphics[width=0.82\linewidth]{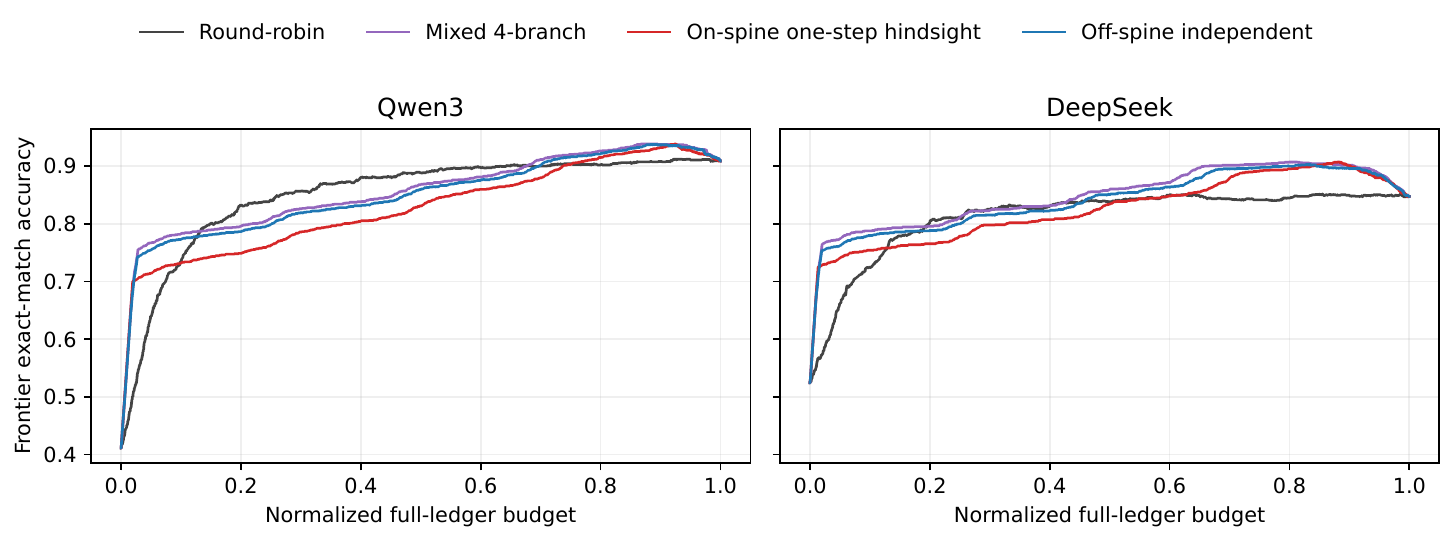}
\caption{Pooled-spine diagnostic separating mixed value into realized
on-spine hindsight and independent off-spine information.  All methods share
the same stored transitions and logical-ledger endpoint.  Accuracy is averaged
over problem--spine pairs.}
\label{fig:oracle-decomposition}
\end{figure*}

\Cref{tab:oracle-decomposition} reports this pooled-spine decomposition; it is
visualized in \Cref{fig:oracle-decomposition}.

\subsection{Cross-spine agreement accounting}

Modal agreement requires four answer probes at the same depth.  We compare
an uncharged diagnostic, which omits the cost of the auxiliary trajectories,
with a charged policy that initializes and advances four synchronized spines.
The charged policy pays for every reasoning path and answer probe; accuracy
is evaluated on the designated spine.

\begin{table*}[t]
\centering
\footnotesize
\resizebox{\linewidth}{!}{%
\begin{tabular}{lllrr}
\toprule
Model & Ledger & Agreement treatment & Mean normalized AULC & Gain vs.\ single-spine RR (95\% CI) \\
\midrule
Qwen3 & Logical tokens & Single-spine round-robin & 0.851 & +0.00\% [+0.00, +0.00] \\
Qwen3 & Logical tokens & Free cross-spine diagnostic & 0.830 & -2.47\% [-4.71, -0.19] \\
Qwen3 & Logical tokens & Charged four-spine policy & 0.627 & -26.29\% [-29.79, -22.92] \\
\addlinespace
Qwen3 & Estimated FLOPs & Single-spine round-robin & 0.854 & +0.00\% [+0.00, +0.00] \\
Qwen3 & Estimated FLOPs & Free cross-spine diagnostic & 0.831 & -2.66\% [-4.78, -0.54] \\
Qwen3 & Estimated FLOPs & Charged four-spine policy & 0.504 & -40.99\% [-45.31, -36.65] \\
\addlinespace
Qwen3 & Calibrated H20 seconds & Single-spine round-robin & 0.854 & +0.00\% [+0.00, +0.00] \\
Qwen3 & Calibrated H20 seconds & Free cross-spine diagnostic & 0.831 & -2.65\% [-4.93, -0.45] \\
Qwen3 & Calibrated H20 seconds & Charged four-spine policy & 0.600 & -29.69\% [-33.40, -25.92] \\
\addlinespace
DeepSeek & Logical tokens & Single-spine round-robin & 0.813 & +0.00\% [+0.00, +0.00] \\
DeepSeek & Logical tokens & Free cross-spine diagnostic & 0.819 & +0.78\% [-0.55, +2.15] \\
DeepSeek & Logical tokens & Charged four-spine policy & 0.672 & -17.28\% [-19.84, -14.67] \\
\addlinespace
DeepSeek & Estimated FLOPs & Single-spine round-robin & 0.814 & +0.00\% [+0.00, +0.00] \\
DeepSeek & Estimated FLOPs & Free cross-spine diagnostic & 0.819 & +0.67\% [-0.63, +2.05] \\
DeepSeek & Estimated FLOPs & Charged four-spine policy & 0.580 & -28.72\% [-32.34, -25.05] \\
\addlinespace
DeepSeek & Calibrated H20 seconds & Single-spine round-robin & 0.814 & +0.00\% [+0.00, +0.00] \\
DeepSeek & Calibrated H20 seconds & Free cross-spine diagnostic & 0.819 & +0.68\% [-0.59, +1.98] \\
DeepSeek & Calibrated H20 seconds & Charged four-spine policy & 0.635 & -22.02\% [-25.02, -18.86] \\
\addlinespace
\bottomrule
\end{tabular}
}
\caption{Cross-spine agreement under different cost ledgers. The free row evaluates agreement without charging its auxiliary trajectories. The charged policy pays the other three reasoning trajectories and answer probes initially and whenever agreement is updated; the problem's designated spine alone defines accuracy. Prompts are deduplicated and their KV state is cloned. Each ledger uses a common endpoint across the three rows.}
\label{tab:agreement-accounting}
\end{table*}

\Cref{tab:agreement-accounting} shows both the free diagnostic and the fully
charged four-spine policy under all three cost coordinates.

\subsection{Exact non-myopic hindsight}

A deterministic problem-ID rule selects 256 distinct confirmation problems per model, partitioned
into 64 non-overlapping groups of four.  Each group contains one 16-state
on-spine chain per problem.  Dynamic programming enumerates all
\(16^4=65{,}536\) precedence-feasible joint states, charges stored transition
cost, and maximizes the same right-continuous AULC objective.  With all transition outcomes known, the optimum gives an exact hindsight
upper bound within each sampled four-problem system.

\begin{table*}[t]
\centering
\footnotesize
\setlength{\tabcolsep}{3.8pt}
\begin{tabular}{llrrr}
\toprule
Model & Policy & Mean normalized AULC & Gain vs.\ RR (95\% CI) & Gap to exact DP (95\% CI) \\
\midrule
Qwen3 & Round-robin & 0.843 & +0.00\% [+0.00, +0.00] & 6.61\% [5.06, 8.31] \\
Qwen3 & Mixed myopic & 0.841 & -0.25\% [-2.41, +1.88] & 6.84\% [4.78, 9.00] \\
Qwen3 & On-spine one-step hindsight & 0.820 & -2.74\% [-5.35, -0.15] & 9.16\% [6.74, 11.67] \\
Qwen3 & Off-spine myopic & 0.832 & -1.22\% [-3.42, +0.81] & 7.75\% [5.65, 10.16] \\
Qwen3 & Exact non-myopic hindsight & 0.902 & +7.08\% [+5.31, +9.03] & 0.00\% [0.00, 0.00] \\
\addlinespace
DeepSeek & Round-robin & 0.766 & +0.00\% [+0.00, +0.00] & 10.36\% [7.86, 13.46] \\
DeepSeek & Mixed myopic & 0.809 & +5.68\% [+2.71, +8.89] & 5.27\% [3.47, 7.23] \\
DeepSeek & On-spine one-step hindsight & 0.791 & +3.31\% [+0.15, +6.58] & 7.39\% [5.08, 9.98] \\
DeepSeek & Off-spine myopic & 0.808 & +5.48\% [+2.77, +8.65] & 5.44\% [3.57, 7.39] \\
DeepSeek & Exact non-myopic hindsight & 0.854 & +11.55\% [+8.19, +15.11] & 0.00\% [0.00, 0.00] \\
\addlinespace
\bottomrule
\end{tabular}
\caption{Exact non-myopic hindsight scheduling on 64 deterministic,
non-overlapping subsets per model.  Each subset contains one complete
16-state frontier from each of four distinct confirmation problems.  Dynamic
programming enumerates all $16^4$ precedence-feasible frontier states and
maximizes full-ledger AULC within each subset.  Intervals bootstrap the 64
four-problem subsets 2,000 times.}
\label{tab:nonmyopic-dp}
\end{table*}

\Cref{tab:nonmyopic-dp} quantifies the gap between one-step orderings and this
subset-specific exact hindsight optimum.

\section{Complete Proxy and Controller Results}
\label{app:proxy-detail}

\subsection{Discrimination, rare-event baselines, and calibration}

The main text uses conditional event-detection AUROC as its primary
prediction endpoint.  \Cref{tab:full-proxy-ci} supplies supporting
discrimination and calibration estimates with problem-bootstrap intervals.
Confidence and ANLL are monotone transformations of the same signal;
confidence represents this pair in the table.  The controllers use both
confidence and ANLL as inputs.

\begin{figure*}[t]
\centering
\includegraphics[width=0.56\linewidth]{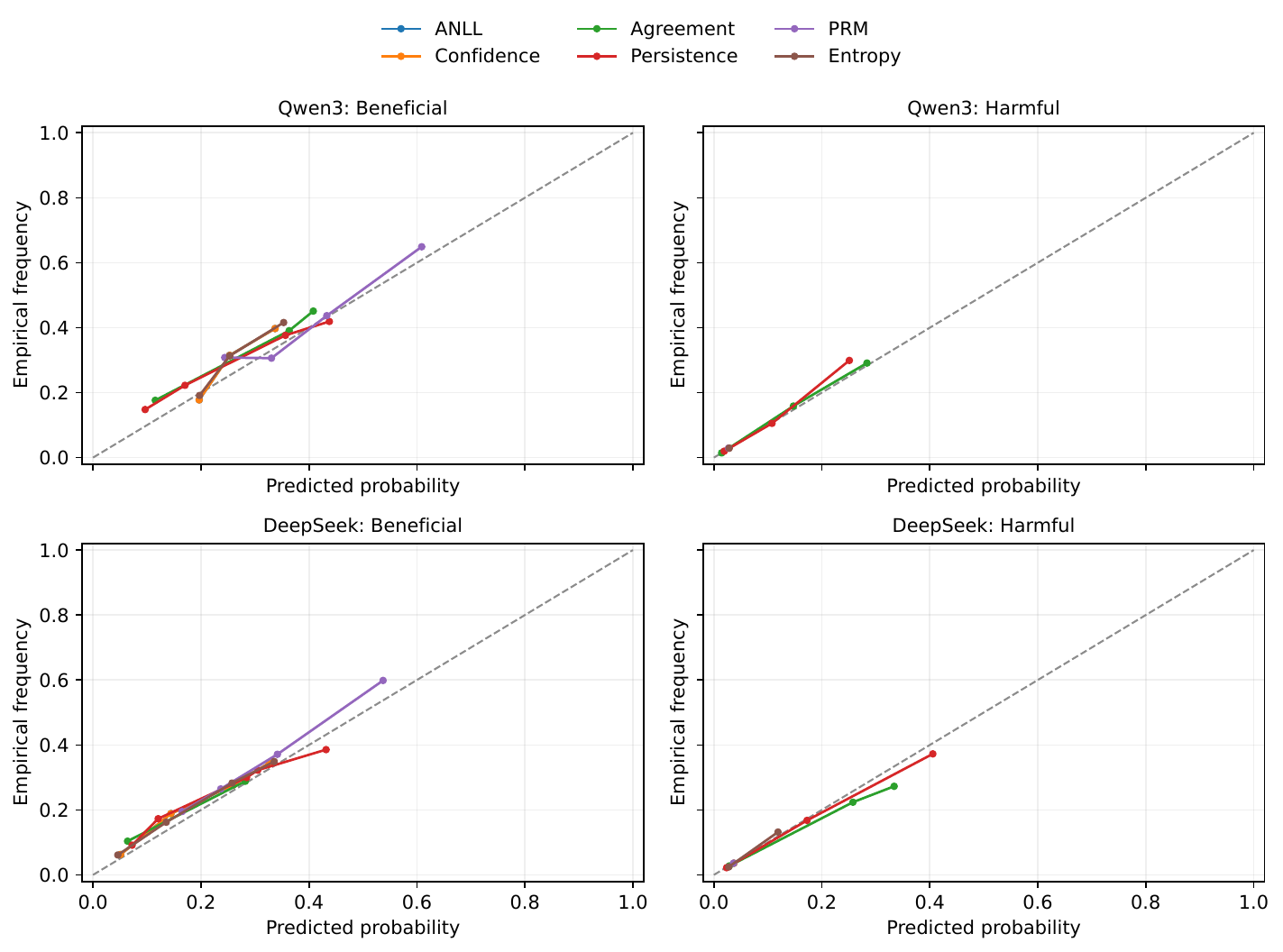}
\caption{Confirmation calibration for all audited proxies.  Orientations and
tie-preserving bins are fixed using analysis problems; the diagonal denotes
perfect calibration.}
\label{fig:calibration}
\end{figure*}

Because damage is rare, \Cref{tab:rare-event-baselines} pairs average
precision and Brier score with their prevalence baselines.  For the best
damage proxies, average precision is 4.34--5.16 times prevalence, while Brier
skill is 0.078--0.099.  This distinguishes useful event ranking from accurate
probability estimation.  \Cref{fig:calibration} shows held-out calibration
curves.

\begin{table*}[t]
\centering
\scriptsize
\setlength{\tabcolsep}{3.4pt}
\begin{tabular}{lllrrrr}
\toprule
Model & Target & Proxy & $n_{\rm conf}$ & AUROC (95\% CI) & AUPRC (95\% CI) & Brier (95\% CI) \\
\midrule
Qwen3 & Beneficial & Confidence & 3,686 & 0.501 [0.428, 0.572] & 0.338 [0.281, 0.415] & 0.227 [0.206, 0.252] \\
Qwen3 & Beneficial & Agreement & 3,678 & 0.625 [0.573, 0.675] & 0.428 [0.366, 0.497] & 0.218 [0.196, 0.242] \\
Qwen3 & Beneficial & Persistence & 2,702 & 0.658 [0.610, 0.706] & 0.363 [0.295, 0.444] & 0.183 [0.155, 0.217] \\
Qwen3 & Beneficial & PRM & 3,686 & 0.604 [0.562, 0.649] & 0.494 [0.432, 0.566] & 0.221 [0.198, 0.246] \\
Qwen3 & Beneficial & Entropy & 3,686 & 0.510 [0.434, 0.584] & 0.334 [0.271, 0.427] & 0.226 [0.204, 0.252] \\
Qwen3 & Harmful & Confidence & 18,634 & 0.598 [0.549, 0.648] & 0.067 [0.042, 0.105] & 0.028 [0.022, 0.034] \\
Qwen3 & Harmful & Agreement & 18,634 & 0.742 [0.699, 0.782] & 0.150 [0.102, 0.207] & 0.025 [0.020, 0.031] \\
Qwen3 & Harmful & Persistence & 18,022 & 0.730 [0.688, 0.772] & 0.111 [0.072, 0.159] & 0.025 [0.020, 0.031] \\
Qwen3 & Harmful & PRM & 18,634 & 0.660 [0.607, 0.707] & 0.056 [0.041, 0.073] & 0.028 [0.022, 0.034] \\
Qwen3 & Harmful & Entropy & 18,634 & 0.623 [0.566, 0.677] & 0.128 [0.071, 0.198] & 0.028 [0.022, 0.034] \\
DeepSeek & Beneficial & Confidence & 4,401 & 0.611 [0.559, 0.659] & 0.307 [0.260, 0.367] & 0.185 [0.166, 0.207] \\
DeepSeek & Beneficial & Agreement & 4,393 & 0.545 [0.493, 0.599] & 0.277 [0.238, 0.328] & 0.188 [0.169, 0.209] \\
DeepSeek & Beneficial & Persistence & 3,665 & 0.673 [0.632, 0.715] & 0.299 [0.247, 0.360] & 0.154 [0.135, 0.178] \\
DeepSeek & Beneficial & PRM & 4,401 & 0.662 [0.620, 0.704] & 0.433 [0.375, 0.504] & 0.178 [0.160, 0.200] \\
DeepSeek & Beneficial & Entropy & 4,401 & 0.611 [0.557, 0.664] & 0.302 [0.258, 0.364] & 0.184 [0.165, 0.205] \\
DeepSeek & Harmful & Confidence & 17,919 & 0.668 [0.630, 0.708] & 0.177 [0.129, 0.233] & 0.034 [0.029, 0.039] \\
DeepSeek & Harmful & Agreement & 17,919 & 0.702 [0.665, 0.737] & 0.112 [0.082, 0.150] & 0.033 [0.028, 0.038] \\
DeepSeek & Harmful & Persistence & 17,139 & 0.716 [0.681, 0.747] & 0.141 [0.106, 0.185] & 0.029 [0.025, 0.034] \\
DeepSeek & Harmful & PRM & 17,919 & 0.584 [0.543, 0.621] & 0.065 [0.051, 0.087] & 0.034 [0.029, 0.040] \\
DeepSeek & Harmful & Entropy & 17,919 & 0.681 [0.636, 0.720] & 0.215 [0.155, 0.277] & 0.034 [0.028, 0.039] \\
\bottomrule
\end{tabular}
\caption{Supporting confirmation discrimination and calibration results with problem-bootstrap intervals. Confidence represents the monotone-equivalent confidence/ANLL pair; separate ANLL results are available in the accompanying results archive.}
\label{tab:full-proxy-ci}
\end{table*}

\begin{table*}[t]
\centering
\scriptsize
\setlength{\tabcolsep}{3.5pt}
\begin{tabular}{lllrrrrrr}
\toprule
Model & Target & Proxy & Analysis prev. & Conf.\ prev. & AP & AP/prev. & Brier ref. & BSS \\
\midrule
Qwen3 & Beneficial & Confidence & 0.307 & 0.357 & 0.338 & 0.95$\times$ & 0.232 & +0.022 \\
Qwen3 & Beneficial & Agreement & 0.306 & 0.356 & 0.428 & 1.20$\times$ & 0.232 & +0.058 \\
Qwen3 & Beneficial & Persistence & 0.224 & 0.264 & 0.363 & 1.37$\times$ & 0.196 & +0.064 \\
Qwen3 & Beneficial & PRM & 0.307 & 0.357 & 0.494 & 1.38$\times$ & 0.232 & +0.050 \\
Qwen3 & Beneficial & Entropy & 0.307 & 0.357 & 0.334 & 0.94$\times$ & 0.232 & +0.025 \\
Qwen3 & Harmful & Confidence & 0.028 & 0.029 & 0.067 & 2.30$\times$ & 0.028 & +0.005 \\
Qwen3 & Harmful & Agreement & 0.028 & 0.029 & 0.150 & 5.16$\times$ & 0.028 & +0.099 \\
Qwen3 & Harmful & Persistence & 0.026 & 0.027 & 0.111 & 4.05$\times$ & 0.027 & +0.064 \\
Qwen3 & Harmful & PRM & 0.028 & 0.029 & 0.056 & 1.91$\times$ & 0.028 & +0.011 \\
Qwen3 & Harmful & Entropy & 0.028 & 0.029 & 0.128 & 4.40$\times$ & 0.028 & +0.009 \\
DeepSeek & Beneficial & Confidence & 0.235 & 0.260 & 0.307 & 1.18$\times$ & 0.193 & +0.040 \\
DeepSeek & Beneficial & Agreement & 0.235 & 0.260 & 0.277 & 1.07$\times$ & 0.193 & +0.026 \\
DeepSeek & Beneficial & Persistence & 0.182 & 0.209 & 0.299 & 1.43$\times$ & 0.166 & +0.070 \\
DeepSeek & Beneficial & PRM & 0.235 & 0.260 & 0.433 & 1.67$\times$ & 0.193 & +0.077 \\
DeepSeek & Beneficial & Entropy & 0.235 & 0.260 & 0.302 & 1.16$\times$ & 0.193 & +0.047 \\
DeepSeek & Harmful & Confidence & 0.036 & 0.036 & 0.177 & 4.93$\times$ & 0.035 & +0.022 \\
DeepSeek & Harmful & Agreement & 0.036 & 0.036 & 0.112 & 3.12$\times$ & 0.035 & +0.057 \\
DeepSeek & Harmful & Persistence & 0.033 & 0.033 & 0.141 & 4.34$\times$ & 0.031 & +0.078 \\
DeepSeek & Harmful & PRM & 0.036 & 0.036 & 0.065 & 1.81$\times$ & 0.035 & +0.008 \\
DeepSeek & Harmful & Entropy & 0.036 & 0.036 & 0.215 & 5.99$\times$ & 0.035 & +0.031 \\
\bottomrule
\end{tabular}
\caption{Rare-event baselines on each proxy's available-case population. Confidence represents the confidence/ANLL pair. Confirmation prevalence is the no-skill AP baseline. The Brier reference predicts the analysis-set prevalence for every confirmation row; $\mathrm{BSS}=1-\mathrm{BS}/\mathrm{BS}_{\rm ref}$.}
\label{tab:rare-event-baselines}
\end{table*}

\subsection{Matched and conditional information}

These diagnostics examine variation in continuation outcomes among
checkpoints with similar observed signals.  Collision coverage is the
fraction of checkpoints in matched cells containing both a feasible nonzero
event and a zero-value control.  Bin sensitivity, exact matching, and
nearest-neighbor comparisons assess this outcome heterogeneity at different
matching resolutions.  Conditional AUROC and held-out log-loss reduction
measure predictive information beyond the matching covariates.

At ten-bin resolution, 81.9\% and 94.5\% of Qwen3 and DeepSeek checkpoints
lie in at least one ambiguous matched cell; exact persistence matching gives
77.3\% and 87.7\%.  Conditional AUROC is 0.632--0.692, and held-out log-loss
reduction is 0.021--0.053 bits.  Thus persistence provides predictive
information even though checkpoints with matching signals can have different
continuation outcomes.

\begin{table*}[!t]
\centering
\begin{minipage}[t]{0.485\textwidth}
\captionsetup{hypcap=false}
\centering
\scriptsize
\renewcommand{\arraystretch}{0.92}
\setlength{\tabcolsep}{1.8pt}
\begin{tabular}{llrrrr}
\toprule
Model & Proxy & Collision & Cells & Recovery gap & Damage gap \\
\midrule
Qwen3 & ANLL & 48.0\% & 606 & +2.703 & -1.813 \\
Qwen3 & Confidence & 48.0\% & 606 & +2.703 & -1.813 \\
Qwen3 & Agreement & 73.7\% & 323 & +2.687 & -1.769 \\
Qwen3 & Persistence & 77.3\% & 327 & +2.616 & -1.777 \\
Qwen3 & PRM & 45.3\% & 591 & +2.677 & -1.806 \\
Qwen3 & Entropy & 47.6\% & 585 & +2.693 & -1.805 \\
DeepSeek & ANLL & 55.8\% & 654 & +2.657 & -1.976 \\
DeepSeek & Confidence & 55.8\% & 654 & +2.657 & -1.976 \\
DeepSeek & Agreement & 87.5\% & 366 & +2.665 & -1.974 \\
DeepSeek & Persistence & 87.7\% & 329 & +2.575 & -1.961 \\
DeepSeek & PRM & 58.2\% & 642 & +2.565 & -2.000 \\
DeepSeek & Entropy & 54.5\% & 619 & +2.662 & -1.949 \\
\bottomrule
\end{tabular}
\captionof{table}{Matched-cell collision diagnostics. Gaps compare feasible
nonzero checkpoints with zero-value controls after matching model, dataset,
difficulty, depth, current correctness, and proxy bin.}
\label{tab:collision-detail}
\end{minipage}
\hfill
\begin{minipage}[t]{0.485\textwidth}
\captionsetup{hypcap=false}
\centering
\scriptsize
\renewcommand{\arraystretch}{0.92}
\begin{tabular}{llrrrr}
\toprule
Model & Proxy & 5 bins & 10 bins & 20 bins & 50 bins \\
\midrule
Qwen3 & ANLL & 63.1\% & 48.0\% & 33.6\% & 20.4\% \\
Qwen3 & Confidence & 63.1\% & 48.0\% & 33.6\% & 20.4\% \\
Qwen3 & Agreement & 73.7\% & 73.7\% & 73.7\% & 73.7\% \\
Qwen3 & Persistence & 77.3\% & 77.3\% & 77.3\% & 77.3\% \\
Qwen3 & PRM & 56.9\% & 45.3\% & 33.1\% & 20.5\% \\
Qwen3 & Entropy & 63.2\% & 47.6\% & 33.2\% & 20.6\% \\
\addlinespace
DeepSeek & ANLL & 74.1\% & 55.8\% & 39.8\% & 23.2\% \\
DeepSeek & Confidence & 74.1\% & 55.8\% & 39.8\% & 23.2\% \\
DeepSeek & Agreement & 87.5\% & 87.5\% & 87.5\% & 87.5\% \\
DeepSeek & Persistence & 87.7\% & 87.7\% & 87.7\% & 87.7\% \\
DeepSeek & PRM & 73.7\% & 58.2\% & 41.9\% & 25.2\% \\
DeepSeek & Entropy & 72.2\% & 54.5\% & 38.1\% & 23.0\% \\
\bottomrule
\end{tabular}
\captionof{table}{Collision sensitivity to proxy resolution. Persistence and
agreement use exact values; continuous-proxy bins are fitted on analysis
data.}
\label{tab:collision-bin-sensitivity}
\end{minipage}
\end{table*}

\begin{table*}[!htbp]
\centering
\scriptsize
\renewcommand{\arraystretch}{0.90}
\begin{tabular}{lllrrr}
\toprule
Model & Target & Proxy & Conditional AUROC & NN disagreement & $\Delta$ test log-loss, bits \\
\midrule
Qwen3 & Beneficial & ANLL & 0.524 & 34.3\% [30.8, 37.9] & -0.0166 [-0.0344, -0.0006] \\
Qwen3 & Beneficial & Confidence & 0.476 & 34.3\% [30.9, 37.7] & -0.0166 [-0.0331, -0.0008] \\
Qwen3 & Beneficial & Agreement & 0.581 & 35.1\% [31.8, 38.7] & +0.0144 [-0.0112, +0.0416] \\
Qwen3 & Beneficial & Persistence & 0.644 & 32.3\% [28.7, 36.5] & +0.0468 [+0.0193, +0.0760] \\
Qwen3 & Beneficial & PRM & 0.597 & 35.6\% [32.9, 38.4] & -0.0085 [-0.0233, +0.0067] \\
Qwen3 & Beneficial & Entropy & 0.534 & 36.7\% [33.4, 40.1] & -0.0239 [-0.0406, -0.0088] \\
\addlinespace
Qwen3 & Harmful & ANLL & 0.562 & 5.5\% [4.7, 6.3] & -0.0046 [-0.0074, -0.0015] \\
Qwen3 & Harmful & Confidence & 0.562 & 5.5\% [4.7, 6.3] & -0.0046 [-0.0074, -0.0016] \\
Qwen3 & Harmful & Agreement & 0.697 & 4.9\% [4.2, 5.7] & +0.0199 [+0.0111, +0.0307] \\
Qwen3 & Harmful & Persistence & 0.691 & 4.7\% [4.1, 5.5] & +0.0215 [+0.0129, +0.0308] \\
Qwen3 & Harmful & PRM & 0.680 & 5.4\% [4.6, 6.2] & +0.0016 [-0.0027, +0.0065] \\
Qwen3 & Harmful & Entropy & 0.587 & 5.1\% [4.4, 5.8] & -0.0039 [-0.0078, +0.0002] \\
\addlinespace
DeepSeek & Beneficial & ANLL & 0.563 & 31.3\% [28.2, 34.6] & -0.0039 [-0.0242, +0.0174] \\
DeepSeek & Beneficial & Confidence & 0.563 & 31.3\% [28.2, 34.6] & -0.0038 [-0.0241, +0.0179] \\
DeepSeek & Beneficial & Agreement & 0.547 & 31.5\% [28.8, 34.3] & +0.0088 [-0.0077, +0.0244] \\
DeepSeek & Beneficial & Persistence & 0.632 & 27.2\% [24.7, 29.8] & +0.0528 [+0.0275, +0.0820] \\
DeepSeek & Beneficial & PRM & 0.621 & 30.3\% [27.9, 33.0] & +0.0188 [-0.0006, +0.0384] \\
DeepSeek & Beneficial & Entropy & 0.573 & 30.7\% [27.9, 33.7] & -0.0000 [-0.0234, +0.0233] \\
\addlinespace
DeepSeek & Harmful & ANLL & 0.629 & 6.2\% [5.6, 6.9] & +0.0058 [-0.0001, +0.0120] \\
DeepSeek & Harmful & Confidence & 0.629 & 6.2\% [5.6, 6.9] & +0.0058 [-0.0000, +0.0121] \\
DeepSeek & Harmful & Agreement & 0.621 & 6.2\% [5.6, 6.9] & +0.0149 [+0.0063, +0.0241] \\
DeepSeek & Harmful & Persistence & 0.634 & 5.5\% [5.0, 6.1] & +0.0242 [+0.0161, +0.0330] \\
DeepSeek & Harmful & PRM & 0.610 & 6.6\% [5.9, 7.3] & -0.0029 [-0.0068, +0.0016] \\
DeepSeek & Harmful & Entropy & 0.635 & 6.1\% [5.5, 6.7] & +0.0065 [+0.0001, +0.0136] \\
\addlinespace
\bottomrule
\end{tabular}
\caption{Conditional and incremental proxy diagnostics on confirmation
problems. Positive $\Delta$ test log-loss favors the proxy-augmented model;
intervals use problem-level bootstrap.}
\label{tab:collision-incremental}
\end{table*}

\Cref{tab:collision-detail,tab:collision-bin-sensitivity,tab:collision-incremental}
respectively report matched-cell ambiguity, bin-resolution sensitivity, and
conditional or incremental information.

Removing pooled difficulty yields coarser matching cells and diagnostic
covariates constructed without confirmation stop labels.  Persistence retains
above-chance conditional AUROC and positive held-out log-loss reduction for
both targets and both models; signed matched gaps retain their feasible
direction.  The relative rankings of individual proxies vary across these
specifications.

\begin{table*}[!htbp]
\centering
\scriptsize
\renewcommand{\arraystretch}{0.92}
\begin{tabular}{llrrrrrr}
\toprule
Model & Proxy & \multicolumn{2}{c}{Collision rate} & \multicolumn{2}{c}{Beneficial $\Delta M_s$} & \multicolumn{2}{c}{Harmful $\Delta M_s$} \\
\cmidrule(lr){3-4}\cmidrule(lr){5-6}\cmidrule(lr){7-8}
 & & $+D$ & $-D$ & $+D$ & $-D$ & $+D$ & $-D$ \\
\midrule
Qwen3 & ANLL & 48.0\% & 81.1\% & +2.703 & +2.735 & -1.813 & -1.815 \\
Qwen3 & Confidence & 48.0\% & 81.1\% & +2.703 & +2.735 & -1.813 & -1.815 \\
Qwen3 & Agreement & 73.7\% & 98.1\% & +2.687 & +2.726 & -1.769 & -1.788 \\
Qwen3 & Persistence & 77.3\% & 99.1\% & +2.616 & +2.627 & -1.777 & -1.795 \\
Qwen3 & PRM & 45.3\% & 67.3\% & +2.677 & +2.699 & -1.806 & -1.814 \\
Qwen3 & Entropy & 47.6\% & 79.1\% & +2.693 & +2.734 & -1.805 & -1.811 \\
\addlinespace
DeepSeek & ANLL & 55.8\% & 82.9\% & +2.657 & +2.717 & -1.976 & -1.989 \\
DeepSeek & Confidence & 55.8\% & 82.9\% & +2.657 & +2.717 & -1.976 & -1.989 \\
DeepSeek & Agreement & 87.5\% & 97.9\% & +2.665 & +2.721 & -1.974 & -1.984 \\
DeepSeek & Persistence & 87.7\% & 99.2\% & +2.575 & +2.596 & -1.961 & -1.982 \\
DeepSeek & PRM & 58.2\% & 82.3\% & +2.565 & +2.602 & -2.000 & -2.000 \\
DeepSeek & Entropy & 54.5\% & 80.6\% & +2.662 & +2.714 & -1.949 & -1.964 \\
\addlinespace
\bottomrule
\end{tabular}
\caption{Matched-collision sensitivity after deleting pooled difficulty.
$+D$ includes difficulty; $-D$ removes it while retaining the other matching
covariates and analysis-fitted proxy bins.}
\label{tab:difficulty-deletion-collision}
\end{table*}

\begin{table*}[!htbp]
\centering
\scriptsize
\renewcommand{\arraystretch}{0.90}
\begin{tabular}{lllrrrrrr}
\toprule
Model & Target & Proxy & \multicolumn{2}{c}{Conditional AUROC} & \multicolumn{2}{c}{NN disagreement} & \multicolumn{2}{c}{$\Delta$ test log-loss, bits} \\
\cmidrule(lr){4-5}\cmidrule(lr){6-7}\cmidrule(lr){8-9}
 & & & $+D$ & $-D$ & $+D$ & $-D$ & $+D$ & $-D$ \\
\midrule
Qwen3 & Beneficial & ANLL & 0.524 & 0.543 & 34.3\% & 36.4\% & -0.0166 & +0.0069 \\
Qwen3 & Beneficial & Confidence & 0.476 & 0.457 & 34.3\% & 36.4\% & -0.0166 & +0.0068 \\
Qwen3 & Beneficial & Agreement & 0.581 & 0.577 & 35.1\% & 36.6\% & +0.0144 & +0.0358 \\
Qwen3 & Beneficial & Persistence & 0.644 & 0.633 & 32.3\% & 33.2\% & +0.0468 & +0.0520 \\
Qwen3 & Beneficial & PRM & 0.597 & 0.609 & 35.6\% & 36.7\% & -0.0085 & +0.0045 \\
Qwen3 & Beneficial & Entropy & 0.534 & 0.552 & 36.7\% & 38.7\% & -0.0239 & +0.0003 \\
\addlinespace
Qwen3 & Harmful & ANLL & 0.562 & 0.542 & 5.5\% & 5.5\% & -0.0046 & -0.0003 \\
Qwen3 & Harmful & Confidence & 0.562 & 0.542 & 5.5\% & 5.5\% & -0.0046 & -0.0003 \\
Qwen3 & Harmful & Agreement & 0.697 & 0.710 & 4.9\% & 5.0\% & +0.0199 & +0.0273 \\
Qwen3 & Harmful & Persistence & 0.691 & 0.718 & 4.7\% & 4.7\% & +0.0215 & +0.0279 \\
Qwen3 & Harmful & PRM & 0.680 & 0.733 & 5.4\% & 5.6\% & +0.0016 & +0.0087 \\
Qwen3 & Harmful & Entropy & 0.587 & 0.562 & 5.1\% & 5.2\% & -0.0039 & +0.0010 \\
\addlinespace
DeepSeek & Beneficial & ANLL & 0.563 & 0.561 & 31.3\% & 31.8\% & -0.0039 & +0.0094 \\
DeepSeek & Beneficial & Confidence & 0.563 & 0.561 & 31.3\% & 31.8\% & -0.0038 & +0.0094 \\
DeepSeek & Beneficial & Agreement & 0.547 & 0.510 & 31.5\% & 32.0\% & +0.0088 & +0.0239 \\
DeepSeek & Beneficial & Persistence & 0.632 & 0.645 & 27.2\% & 28.0\% & +0.0528 & +0.0618 \\
DeepSeek & Beneficial & PRM & 0.621 & 0.640 & 30.3\% & 31.1\% & +0.0188 & +0.0376 \\
DeepSeek & Beneficial & Entropy & 0.573 & 0.566 & 30.7\% & 32.0\% & -0.0000 & +0.0109 \\
\addlinespace
DeepSeek & Harmful & ANLL & 0.629 & 0.631 & 6.2\% & 6.2\% & +0.0058 & +0.0098 \\
DeepSeek & Harmful & Confidence & 0.629 & 0.631 & 6.2\% & 6.2\% & +0.0058 & +0.0098 \\
DeepSeek & Harmful & Agreement & 0.621 & 0.649 & 6.2\% & 6.3\% & +0.0149 & +0.0216 \\
DeepSeek & Harmful & Persistence & 0.634 & 0.654 & 5.5\% & 5.6\% & +0.0242 & +0.0287 \\
DeepSeek & Harmful & PRM & 0.610 & 0.643 & 6.6\% & 6.7\% & -0.0029 & +0.0033 \\
DeepSeek & Harmful & Entropy & 0.635 & 0.639 & 6.1\% & 6.2\% & +0.0065 & +0.0107 \\
\addlinespace
\bottomrule
\end{tabular}
\caption{Local-information sensitivity after deleting pooled difficulty.
$+D$ includes and $-D$ excludes difficulty from conditional AUROC,
nearest-neighbor matching, and held-out log-loss models.}
\label{tab:difficulty-deletion-information}
\end{table*}

\Cref{tab:difficulty-deletion-collision,tab:difficulty-deletion-information}
show that the main qualitative conclusions survive deletion of the
label-aware difficulty covariate.

\subsection{Predictor decomposition}

\begin{table*}[t]
\centering
\footnotesize
\renewcommand{\arraystretch}{0.96}
\begin{tabular}{llrrr}
\toprule
Model & Controller & Pearson $r$ & Beneficial AUROC & \shortstack{Harmful AUROC\\($-\widehat M_s$ score)} \\
\midrule
Qwen3 & Ridge-$M$ (single-spine) & 0.373 & 0.715 & 0.291 \\
Qwen3 & Ridge-$M$ (+ charged PRM) & 0.374 & 0.715 & 0.310 \\
DeepSeek & Ridge-$M$ (single-spine) & 0.282 & 0.726 & 0.226 \\
DeepSeek & Ridge-$M$ (+ charged PRM) & 0.284 & 0.730 & 0.235 \\
\bottomrule
\end{tabular}
\caption{Held-out Ridge-$M$ diagnostics. Predictors exclude correctness,
difficulty, agreement, and continuation outcomes; harmful AUROC uses
\(-\widehat M_s\). PRM scoring is charged.}
\label{tab:learned-controller}
\end{table*}

Direct Ridge-\(M\) predicts signed value from persistence, confidence, ANLL,
entropy, depth, and dataset; it excludes current correctness, difficulty,
agreement, and continuation outcomes.  Harmful AUROC uses
\(-\widehat M_s\).  The two-stage model separately predicts current wrongness,
recovery, and damage, then applies analysis-set costs.
\Cref{tab:learned-controller} reports held-out Ridge-\(M\) discrimination,
clarifying why event ranking need not improve allocation.

\section{Intervention Robustness}
\label{app:intervention-robustness}

\subsection{Dataset and depth trajectories}
\begin{table*}[t]
\centering
\footnotesize
\begin{tabular}{llrrrr}
\toprule
Model & Data & Wrong $n$ & Beneficial rate (95\% CI) & Correct $n$ & Harmful rate (95\% CI) \\
\midrule
\addlinespace
Qwen3 & All & 3,686 & 35.7\% [30.4, 42.1] & 18,634 & 2.9\% [2.3, 3.6] \\
Qwen3 & GSM8K & 1,481 & 54.0\% [45.9, 64.3] & 14,479 & 1.8\% [1.3, 2.5] \\
Qwen3 & MATH-500 & 2,205 & 23.4\% [17.7, 30.6] & 4,155 & 6.7\% [4.7, 9.2] \\
\addlinespace
DeepSeek & All & 4,401 & 26.0\% [22.2, 30.4] & 17,919 & 3.6\% [3.0, 4.2] \\
DeepSeek & GSM8K & 1,985 & 32.2\% [26.4, 39.4] & 13,975 & 2.9\% [2.4, 3.5] \\
DeepSeek & MATH-500 & 2,416 & 20.9\% [16.3, 26.4] & 3,944 & 6.0\% [4.1, 8.3] \\
\bottomrule
\end{tabular}
\caption{Confirmation prevalence of signed continuation events. Intervals resample complete problems 2,000 times.}
\label{tab:rq1-prevalence}
\end{table*}

\Cref{tab:rq1-prevalence} gives the model--dataset prevalence, magnitude, and
depth-trend summaries behind the main-text trajectories.

\subsection{Persistence ties, branch count, and ratio definition}

Persistence is discrete and heavily concentrated at one.  The primary
high-first rule compares exact persistence \(p=1\) with all lower values,
rather than splitting tied observations.  Stricter low-persistence thresholds
strengthen the same direction.

\begin{table*}[t]
\centering
\footnotesize
\begin{tabular}{llrrr}
\toprule
Model & Low group & $n_{\rm high}$ & $n_{\rm low}$ & High--low $\Delta M_s$ (95\% CI) \\
\midrule
Qwen3 & $p\leq0.75$ (tie-resolved) & 15,349 & 2,673 & +0.200 [+0.135, +0.276] \\
Qwen3 & $p\leq0.50$ & 15,349 & 1,251 & +0.291 [+0.190, +0.401] \\
Qwen3 & $p\leq0.33$ & 15,349 & 398 & +0.450 [+0.267, +0.665] \\
DeepSeek & $p\leq0.75$ (tie-resolved) & 15,011 & 2,128 & +0.277 [+0.201, +0.361] \\
DeepSeek & $p\leq0.50$ & 15,011 & 937 & +0.480 [+0.341, +0.621] \\
DeepSeek & $p\leq0.33$ & 15,011 & 311 & +0.740 [+0.536, +0.965] \\
\bottomrule
\end{tabular}
\caption{Persistence-threshold sensitivity among currently correct confirmation checkpoints. Both analysis quartiles equal 1.0 because persistence is discrete; the implemented tie order therefore compares $p=1$ with $p<1$. Tighter low groups leave the conclusion unchanged.}
\label{tab:persistence-threshold}
\end{table*}

\Cref{tab:persistence-threshold} reports the value contrasts under these
persistence thresholds.

To measure branch-count sensitivity, \Cref{tab:k-sensitivity} averages
detection of a correctness change over all \(\binom4K\) subsets of the observed
branches.  With one draw, this measures the probability of a change after one
continuation; with multiple draws, it measures the opportunity detected
across the sampled continuations.

Increasing from one to four draws raises recovery detection from 26.6\% to
35.7\% for Qwen3 and from 18.9\% to 26.0\% for DeepSeek.  Damage detection
rises from 1.4\% to 2.9\% and from 1.9\% to 3.6\%, respectively.
Changing the cost-ratio definition has a much smaller effect: replacing the
ratio of expectations with the mean branch-specific ratio preserves every
event sign, has Spearman correlation 0.99999, and changes magnitude by
\(2\times10^{-4}\) on average.

\begin{table*}[t]
\centering
\footnotesize
\begin{tabular}{llrrr}
\toprule
Model & Target & $K$ & Expected event rate & Agreement with $K=4$ \\
\midrule
Qwen3 & Beneficial & 1 & 26.6\% & 90.9\% \\
Qwen3 & Beneficial & 2 & 31.2\% & 95.4\% \\
Qwen3 & Beneficial & 3 & 33.8\% & 98.1\% \\
Qwen3 & Beneficial & 4 & 35.7\% & 100.0\% \\
Qwen3 & Harmful & 1 & 1.4\% & 98.5\% \\
Qwen3 & Harmful & 2 & 2.0\% & 99.1\% \\
Qwen3 & Harmful & 3 & 2.5\% & 99.6\% \\
Qwen3 & Harmful & 4 & 2.9\% & 100.0\% \\
DeepSeek & Beneficial & 1 & 18.9\% & 92.9\% \\
DeepSeek & Beneficial & 2 & 22.5\% & 96.5\% \\
DeepSeek & Beneficial & 3 & 24.5\% & 98.5\% \\
DeepSeek & Beneficial & 4 & 26.0\% & 100.0\% \\
DeepSeek & Harmful & 1 & 1.9\% & 98.3\% \\
DeepSeek & Harmful & 2 & 2.6\% & 99.0\% \\
DeepSeek & Harmful & 3 & 3.2\% & 99.6\% \\
DeepSeek & Harmful & 4 & 3.6\% & 100.0\% \\
\bottomrule
\end{tabular}
\caption{Branch-count sensitivity on confirmation problems. For $K<4$, values average over every size-$K$ subset of the four observed branches, avoiding dependence on the distinguished on-spine branch order.}
\label{tab:k-sensitivity}
\end{table*}

\begin{table*}[t]
\centering
\footnotesize
\begin{tabular}{lrrrr}
\toprule
Model & Pearson & Spearman & Sign agree. & Mean $|\Delta|$ \\
\midrule
Qwen3 & 1.00000 & 0.99999 & 100.0\% & 0.0002 \\
DeepSeek & 1.00000 & 0.99999 & 100.0\% & 0.0002 \\
\bottomrule
\end{tabular}
\caption{Sensitivity of the signed value estimand on confirmation problems. The primary ratio of expected correctness change to expected logical cost is compared with the mean of four branch-specific change-per-cost ratios. Both beneficial and harmful event labels agree exactly in both models; correlations quantify the small remaining magnitude and ranking differences.}
\label{tab:value-ratio-sensitivity}
\end{table*}

\Cref{tab:k-sensitivity,tab:value-ratio-sensitivity} summarize the effects
of branch count and cost-ratio definition.

\subsection{Continuation length}
\label{sec:block-length-sensitivity}

A deterministic problem-ID rule selects 400 problems paired across models and balanced between
analysis and confirmation (150 GSM8K and 50 MATH-500 per split).  All four
spines and the common depth lattice through 3,584 tokens are retained.  The
128-token arm truncates the stored branch, the 256-token arm reuses it, and the
512-token arm extends the same branch with a second contiguous 256-token block
conditioned on the first.  Generation remains label-blind.

For each model and \(\ell\in\{128,256,512\}\), an analysis-only Ridge model
predicts signed change per 1,000 logical tokens from dataset and checkpoint depth:
\[
 \ell^\star(s)=\arg\max_{\ell}\widehat V_\ell
   (\text{dataset},\text{depth}),
\]
with ties resolved toward the shorter length.  This context-only rule assigns
the same length to every checkpoint in a model--dataset--depth cell.

\begin{table*}[t]
\centering
\footnotesize
\begin{tabular}{lrrrr}
\toprule
Length & Recovery & Damage & Mean \(M_s^{\mathcal A}\) &
\(\Delta M\) vs.\ 256 (95\% CI) \\
\midrule
\multicolumn{5}{l}{\textit{Qwen3}} \\
128 tokens & 19.0\% [15.0, 24.1] & 1.6\% [1.1, 2.2] &
0.137 & +0.016 [-0.001, +0.033] \\
256 tokens & 26.6\% [21.2, 33.6] & 1.5\% [1.0, 2.1] &
0.121 & +0.000 [+0.000, +0.000] \\
512 tokens & 36.0\% [28.9, 45.5] & 1.6\% [1.0, 2.3] &
0.090 & -0.031 [-0.040, -0.023] \\
Adaptive & 26.8\% [21.2, 33.7] & 1.6\% [1.0, 2.2] &
0.121 & +0.000 [-0.008, +0.007] \\
\addlinespace
\multicolumn{5}{l}{\textit{DeepSeek}} \\
128 tokens & 14.3\% [11.4, 18.0] & 1.7\% [1.3, 2.2] &
0.113 & +0.021 [+0.002, +0.039] \\
256 tokens & 19.6\% [16.2, 24.3] & 1.9\% [1.5, 2.5] &
0.092 & +0.000 [+0.000, +0.000] \\
512 tokens & 25.5\% [20.6, 31.9] & 2.0\% [1.5, 2.6] &
0.068 & -0.023 [-0.031, -0.016] \\
Adaptive & 14.9\% [12.1, 18.7] & 1.9\% [1.4, 2.4] &
0.115 & \keyresult{+0.023 [+0.007, +0.039]} \\
\bottomrule
\end{tabular}
\caption{Complete continuation-length sensitivity on the 200-problem
confirmation ablation.  Recovery and damage are single-draw rates; \(\Delta M\)
compares four-branch value per 1,000 logical tokens with fixed 256.}
\label{tab:block-length-events-full}
\end{table*}

\begin{table*}[t]
\centering
\footnotesize
\begin{tabular}{lllll}
\toprule
Model & Target & Condition & Rank $\rho$ vs.\ 256 & Top proxy (half-sample stability) \\
\midrule
Qwen3 & Recovery & 128 tokens & 1.00 [0.82, 1.00] & persistence (89\%) \\
Qwen3 & Recovery & 512 tokens & 1.00 [-0.22, 1.00] & persistence (80\%) \\
Qwen3 & Recovery & Adaptive & 1.00 [0.82, 1.00] & persistence (95\%) \\
Qwen3 & Damage & 128 tokens & 1.00 [0.76, 1.00] & modal\_agreement (88\%) \\
Qwen3 & Damage & 512 tokens & 1.00 [0.76, 1.00] & modal\_agreement (80\%) \\
Qwen3 & Damage & Adaptive & 1.00 [0.82, 1.00] & modal\_agreement (92\%) \\
DeepSeek & Recovery & 128 tokens & 1.00 [0.65, 1.00] & persistence (88\%) \\
DeepSeek & Recovery & 512 tokens & 1.00 [0.55, 1.00] & persistence (84\%) \\
DeepSeek & Recovery & Adaptive & 1.00 [0.65, 1.00] & persistence (86\%) \\
DeepSeek & Damage & 128 tokens & 0.82 [0.09, 1.00] & top20\_entropy (45\%) \\
DeepSeek & Damage & 512 tokens & 0.94 [0.19, 1.00] & modal\_agreement (62\%) \\
DeepSeek & Damage & Adaptive & 0.94 [0.19, 1.00] & top20\_entropy (44\%) \\
\bottomrule
\end{tabular}
\caption{Stability of the six-proxy AUROC ranking.  Proxy orientations are
selected on the 256-token analysis arm and fixed across confirmation
conditions.  The adaptive selector uses dataset and depth.  Brackets and
parentheses summarize variation over 64 deterministic paired half-problem
subsamples; these are robustness ranges rather than confidence intervals.}
\label{tab:block-length-proxies}
\end{table*}

\Cref{tab:block-length-events-full,tab:block-length-proxies} report the full
paired event-rate intervals and proxy-ranking sensitivity.

Recovery rises with length in both models, while every paired damage contrast
with 256 tokens includes zero.  Persistence ranks first for recovery at every
length in the full sample.  Rankings vary more across half-sample evaluations,
especially for 512-token recovery and rare DeepSeek damage.

Longer blocks expose more repairs but yield lower cost-normalized value.
Mean signed change per 1,000 logical tokens falls from 0.137/0.113 at 128
tokens to 0.090/0.068 at 512 for Qwen3/DeepSeek.  The adaptive cell selector matches fixed 256 on Qwen3 and
improves value by \(+0.023\) on DeepSeek.  The predeclared 256-token action lies between shorter, cheaper continuations
and longer continuations that expose more repairs.

\subsection{Answer cue and parser}

A deterministic 200-problem confirmation sample reuses every frozen prefix
and continuation while replacing only the greedy boxed-answer suffix.  The
canonical rerun quantifies execution reproducibility; all contexts remain in
the analysis whether or not tokens replay exactly.

\begin{table*}[t]
\centering
\footnotesize
\begin{tabular}{lccc}
\toprule
Model & Token identity & Raw-text identity & Normalized-answer identity \\
\midrule
Qwen3 & 98.9\% [98.7, 99.1] & 99.6\% [99.4, 99.7] & 98.5\% [97.5, 99.2] \\
DeepSeek & 98.3\% [98.1, 98.5] & 99.5\% [99.4, 99.7] & 98.8\% [98.0, 99.5] \\
\bottomrule
\end{tabular}

\caption{Canonical-control replay against the primary outputs.}
\label{tab:probe-replay}
\end{table*}

\begin{table*}[t]
\centering
\footnotesize
\setlength{\tabcolsep}{3pt}
\begin{tabular}{llcccc}
\toprule
Model & Cue & Parse rate & Answer agreement & Benefit agreement & Harm agreement \\
\midrule
Qwen3 & Canonical & 99.4\% [98.9, 99.7] & 100.0\% [100.0, 100.0] & 100.0\% [100.0, 100.0] & 100.0\% [100.0, 100.0] \\
Qwen3 & Direct & 51.7\% [47.9, 55.7] & 78.8\% [73.0, 84.5] & 72.0\% [64.0, 78.3] & 80.3\% [79.1, 81.5] \\
Qwen3 & Terse & 99.9\% [99.8, 99.9] & 95.2\% [93.4, 96.7] & 91.0\% [87.2, 94.4] & 96.5\% [95.7, 97.2] \\
DeepSeek & Canonical & 99.8\% [99.4, 100.0] & 100.0\% [100.0, 100.0] & 100.0\% [100.0, 100.0] & 100.0\% [100.0, 100.0] \\
DeepSeek & Direct & 80.8\% [78.1, 83.3] & 87.0\% [83.0, 90.7] & 85.9\% [81.5, 89.5] & 83.4\% [82.0, 84.7] \\
DeepSeek & Terse & 99.8\% [99.5, 100.0] & 95.3\% [93.9, 96.5] & 89.8\% [86.0, 93.1] & 97.0\% [96.5, 97.6] \\
\bottomrule
\end{tabular}

\caption{Answer-cue sensitivity on the deterministic confirmation sample.
Event agreement conditions on the canonical stop population.  Unparseable
alternative endpoints are scored as incorrect.}
\label{tab:probe-sensitivity}
\end{table*}

\Cref{tab:probe-replay,tab:probe-sensitivity} report execution reproducibility
under the canonical cue and sensitivity to alternative cues.  The terse cue
has parse rates of 99.9\% for Qwen3 and 99.8\% for DeepSeek and preserves
89.8--91.0\% of recovery labels and 96.5--97.0\% of damage labels.  The direct
cue parses 51.7\% and 80.8\%, respectively, and changes the measured
continuation effects.  These differences show how the answer procedure
shapes the audit target.

For an independent automated check, a deterministic problem-ID rule selects
up to 200 confirmation transitions per model--dataset--primary-label cell.
A secondary evaluator re-extracts boxed answers and applies exact rational
comparison on GSM8K and symbolic equivalence on MATH-500.  Agreement with the
primary transition label ranges from 95.5\% to 100\% among resolved cases.
We report resolved-cell coverage alongside agreement to distinguish the
scope of this automated check from its agreement rate.

\raggedbottom
\Cref{tab:parser-adjudication} reports the stratified secondary-adjudication
agreement and resolved-cell coverage.

\begin{table*}[t]
\begin{minipage}[t]{\columnwidth}
\vspace{0pt}
\begin{center}
\begin{minipage}{\columnwidth}
\centering
\scriptsize
\setlength{\tabcolsep}{2.6pt}
\begin{tabular}{@{}llrrl@{}}
\toprule
Data & Transition & Population & Resolved & Agreement (95\% CI) \\
\midrule
\multicolumn{5}{l}{\textit{DeepSeek}} \\
GSM8K & beneficial & 1,983 & 199 & 99.0\% [96.7, 100.0] \\
GSM8K & harmful & 767 & 196 & 100.0\% [100.0, 100.0] \\
GSM8K & zero correct & 55,133 & 200 & 100.0\% [100.0, 100.0] \\
GSM8K & zero wrong & 5,957 & 191 & 100.0\% [100.0, 100.0] \\
MATH-500 & beneficial & 1,337 & 200 & 98.5\% [95.7, 100.0] \\
MATH-500 & harmful & 570 & 200 & 98.5\% [95.8, 100.0] \\
MATH-500 & zero correct & 15,206 & 200 & 100.0\% [100.0, 100.0] \\
MATH-500 & zero wrong & 8,327 & 200 & 95.5\% [88.5, 100.0] \\
\addlinespace
\multicolumn{5}{l}{\textit{Qwen3}} \\
GSM8K & beneficial & 2,484 & 172 & 100.0\% [100.0, 100.0] \\
GSM8K & harmful & 455 & 165 & 100.0\% [100.0, 100.0] \\
GSM8K & zero correct & 57,461 & 200 & 100.0\% [100.0, 100.0] \\
GSM8K & zero wrong & 3,440 & 164 & 100.0\% [100.0, 100.0] \\
MATH-500 & beneficial & 1,436 & 200 & 99.0\% [96.7, 100.0] \\
MATH-500 & harmful & 569 & 200 & 99.0\% [97.0, 100.0] \\
MATH-500 & zero correct & 16,051 & 200 & 100.0\% [100.0, 100.0] \\
MATH-500 & zero wrong & 7,384 & 200 & 97.0\% [90.2, 100.0] \\
\bottomrule
\end{tabular}
\captionof{table}{Stratified automated secondary adjudication. Agreement is measured
over resolved cases; intervals use 2,000 problem-cluster bootstrap
replicates.}
\label{tab:parser-adjudication}
\end{minipage}
\end{center}

\begin{center}
\begin{minipage}{\columnwidth}
\centering
\captionsetup{hypcap=false}
\fontsize{8.5}{10}\selectfont
\setlength{\tabcolsep}{2.5pt}
\begin{tabular}{@{}llccc@{}}
\toprule
Target & Best proxy & AUROC & AUPRC & Brier \\
\midrule
Beneficial & persistence & \shortstack{0.729\\{}[0.695, 0.763]} & 0.216 & 0.089 \\
Harmful & persistence & \shortstack{0.730\\{}[0.687, 0.768]} & 0.090 & 0.018 \\
\bottomrule
\end{tabular}
\captionof{table}{Best-proxy results in the DeepSeek-R1-Distill-Llama-8B
replication. Difficulty strata exclude Llama outcomes.}
\label{tab:llama-replication}
\end{minipage}
\end{center}

\end{minipage}\hfill
\begin{minipage}[t]{\columnwidth}
\vspace{0pt}
\begin{center}
\begin{minipage}{\columnwidth}
\centering
\captionsetup{hypcap=false}
\fontsize{8.5}{10}\selectfont
\setlength{\tabcolsep}{3.2pt}
\begin{tabular}{@{}lrrr@{}}
\toprule
Model & Gen. output & Gen. GPU-h & Gen. tok/s \\
\midrule
Qwen3 & 115.15M & 71.8 & 445.8 \\
DeepSeek & 115.19M & 61.7 & 518.5 \\
\midrule
Model & Peak GiB & PRM GPU-h & Total GPU-h \\
\midrule
Qwen3 & 85.5 & 2.7 & 75.1 \\
DeepSeek & 85.5 & 2.8 & 65.1 \\
\bottomrule
\end{tabular}
\captionof{table}{Physical compute over all 32 H20 workers. Total GPU-hours include
generation, PRM scoring, and recorded model loading.}
\label{tab:physical-ledger}
\end{minipage}
\end{center}

\begin{center}
\begin{minipage}{\columnwidth}
\centering
\captionsetup{hypcap=false}
\fontsize{8.5}{10}\selectfont
\setlength{\tabcolsep}{2.4pt}
\begin{tabular}{@{}lrrc@{}}
\toprule
Ledger & RR AULC & PRM AULC & Gain vs.\ RR \\
\midrule
\multicolumn{4}{@{}l}{\textit{Qwen3}} \\
Tokens & 0.851 & 0.439 & \shortstack{-48.41\%\\{}[-52.94, -43.76]} \\
FLOPs & 0.854 & 0.448 & \shortstack{-47.55\%\\{}[-52.20, -43.22]} \\
H20 seconds & 0.854 & 0.567 & \shortstack{-33.53\%\\{}[-37.35, -29.54]} \\
\addlinespace
\multicolumn{4}{@{}l}{\textit{DeepSeek}} \\
Tokens & 0.813 & 0.537 & \shortstack{-33.88\%\\{}[-37.99, -29.68]} \\
FLOPs & 0.814 & 0.539 & \shortstack{-33.71\%\\{}[-37.82, -29.70]} \\
H20 seconds & 0.814 & 0.609 & \shortstack{-25.18\%\\{}[-28.64, -21.75]} \\
\bottomrule
\end{tabular}
\captionof{table}{Cost-ledger sensitivity of PRM scheduling. Each coordinate uses its
own common replay endpoint; H20 seconds are amortized compute estimates rather
than online latency. Brackets are 95\% intervals.}
\label{tab:multi-ledger-prm}
\end{minipage}
\end{center}

\end{minipage}
\end{table*}
\section{External Replication and Cost Sensitivity}
\label{app:external-replication}

\subsection{Full-protocol Llama replication}

DeepSeek-R1-Distill-Llama-8B repeats the complete prefix lattice and
continuation intervention on all 1,819 problems
~\citep{deepseekai2025r1,grattafiori2024llama3}.  Difficulty remains frozen
from the two primary models, so no Llama outcome defines its strata.

\Cref{tab:llama-replication} complements the main-text Llama estimand and
zero-refit transfer tables with complete best-proxy metrics.

The transfer experiment fits the correctness shortcut, two-stage EV, and
Ridge-\(M\) on analysis rows from the two primary models, then freezes feature
order, imputation, scaling, missingness indicators, regularization,
coefficients, and dataset--depth continuation moments.  No Llama row is used
for fitting or hyperparameter selection.

\subsection{Physical compute and alternative ledgers}
\label{app:cost-sensitivity}

\Cref{tab:physical-ledger,tab:multi-ledger-prm} report the measured compute
sidecar and replay sensitivity across logical tokens, estimated FLOPs, and
calibrated H20 seconds.

The generator sidecar records 71.8 Qwen3 and 61.7 DeepSeek generation
GPU-hours, plus 2.7 and 2.8 PRM GPU-hours.  Including recorded model-load time
gives 75.1 and 65.1 GPU-hours.  PRM ordering remains below the compared
single-spine baselines under logical tokens, estimated FLOPs, and calibrated
H20 seconds, although the magnitudes differ across coordinates.

\hyphenation{order outcomes problem relative reproduces}
\section{Exploratory Analyses of Model Differences and Recovery}
\label{app:cpu-analysis}

These exploratory analyses examine how dataset composition, score
calibration, and cost weighting affect Ridge-\(M\)'s gains, and characterize
recovery patterns in stored trajectories.  We use recorded
generations and frozen predictors fitted only on the analysis split.
Intervals are unadjusted 95\% percentile intervals from 2,000 problem-level
bootstrap draws, keeping each problem's trajectories and checkpoints together.

\subsection{Dataset composition and Ridge-\(M\) gains}
\label{app:cpu-dataset-replay}

\paragraph{Contributions to pooled replay.}
We decompose the 372-problem replay in
\Cref{tab:controller-decomposition-replay} by dataset, keeping each policy's
global action order fixed.  A dataset's contribution is the sum of its
per-problem AULC changes divided by 372, averaged over four spines.  The two
dataset contributions therefore sum to the pooled absolute change.
For DeepSeek, MATH-500 contributes \(+0.7821\) percentage points to the
normalized-AULC change from the correctness shortcut to Ridge-\(M\), and
GSM8K contributes \(-0.2072\), totaling \(+0.5749\).  The corresponding
Qwen3 contributions are \(-0.0422\) and \(-0.1356\), totaling \(-0.1778\)
points.  These contributions locate the gains and losses within the pooled
schedule.

\paragraph{Allocation within each dataset.}
We separately replay the pooled, GSM8K, and MATH-500 problem sets using the
paired 200-problem confirmation subset from the continuation-length
ablation: 150 GSM8K and 50 MATH-500 problems per model, four stored spines,
and 14 checkpoints at depths 256--3,584.  This evaluation uses a smaller pool
and shorter horizon than the 372-problem, 16-checkpoint replay above.
The primary audit's answer-now labels determine frontier accuracy.  Initial
prompt, reasoning, and answer-probe costs and each subsequent
reasoning-plus-probe cost enter the logical-token ledger.  AULC is normalized
over the budget interval from the initial observable frontier to the common
endpoint.

Each bootstrap draw uses the same problem sample across models and policies
and reruns global scheduling within the selected pool.  Duplicate draws
become separate problem instances, each retaining its stored trajectories.
We average AULC over the four spines before computing relative gain as
\(100(A_{\mathrm{Ridge}}/A_{\mathrm{shortcut}}-1)\).

\begin{table}[t]
\centering
\small
\setlength{\tabcolsep}{3pt}
\begin{tabular}{lrrl}
\toprule
Pool & \(N\) & Gain (\%) & 95\% CI \\
\midrule
\multicolumn{4}{l}{\textbf{Qwen3}} \\
Pooled & 200 & \(-0.08\) & \([-0.44, 0.32]\) \\
GSM8K & 150 & \(-0.19\) & \([-0.44, 0.05]\) \\
MATH-500 & 50 & \(+0.94\) & \([-1.21, 3.12]\) \\
\midrule
\multicolumn{4}{l}{\textbf{DeepSeek}} \\
Pooled & 200 & \(+0.82\) & \([-0.09, 1.95]\) \\
GSM8K & 150 & \(-0.46\) & \([-0.74, -0.17]\) \\
MATH-500 & 50 & \(+4.09\) & \([0.76, 7.97]\) \\
\bottomrule
\end{tabular}
\caption{Dataset-specific replay on the paired, truncated confirmation subset. Gain is relative to the correctness shortcut. Each bootstrap replicate reruns scheduling with frozen predictors and trajectories; \(N\) counts problems.}
\label{tab:cpu-replay}
\end{table}

\Cref{tab:cpu-replay} shows that DeepSeek's benefit is concentrated on
MATH-500, whereas its GSM8K allocation is worse than the shortcut.
Both models' pooled gain intervals include zero.  The pooled
DeepSeek-minus-Qwen3 difference in relative gain is \(+0.8955\) percentage
points, with interval \([0.0050,1.9923]\).  The lower bound is close to zero,
and the cross-model interval within each dataset includes zero.  The evidence
thus supports task-dependent gains, with limited support for a model-wide
advantage.

\subsection{Score calibration and cost weighting}
\label{app:cpu-control-decomposition}

We use the paired 200-problem confirmation subset in
\Cref{app:cpu-dataset-replay}.  For each model, calibration uses 200 analysis
problems (150 GSM8K and 50 MATH-500), disjoint from confirmation.  Paired
bootstrap draws keep fitted predictors, calibrators, and trajectories fixed.

\paragraph{Calibration and cross-dataset ordering.}
For Ridge-\(M\) and the shortcut separately, we fit a positive-slope
affine map on each dataset by least squares against the original four-branch
value in Eq.~\ref{eq:value}.  The slope floor of \(10^{-6}\) is inactive in
all fits.  These maps preserve within-dataset ranking and change the
interleaving of actions across datasets.  A global positive-slope map
preserves the entire schedule and gives identical AULC.

\begin{table*}[t]
\centering
\small
\setlength{\tabcolsep}{5pt}
\begin{tabular}{lccc}
\toprule
Model & \shortstack{Raw\\gain (\%)} & \shortstack{Calibrated\\gain (\%)} &
\shortstack{Calibration effect\\on AULC advantage (pp)} \\
\midrule
Qwen3 & \(-0.077\;[-0.439, +0.323]\) & \(+0.263\;[-0.471, +1.130]\) & \(+0.293\;[-0.160, +0.807]\) \\
DeepSeek & \(+0.818\;[-0.094, +1.955]\) & \(+0.719\;[-0.194, +1.880]\) & \(-0.079\;[-0.459, +0.278]\) \\
\bottomrule
\end{tabular}
\caption{Dataset-wise calibration on the paired 200-problem confirmation
subset. Gains compare Ridge-\(M\) with the shortcut under the same calibration
rule. The last column gives the calibrated-minus-raw change in their AULC
difference, in percentage points (pp). Brackets give paired 95\%
problem-bootstrap intervals.}
\label{tab:cpu-control-calibration}
\end{table*}

\Cref{tab:cpu-control-calibration} reports gains against the corresponding
shortcut.  The DeepSeek-minus-Qwen3 relative-gain gap changes by
\(-0.4401\) percentage points, with interval \([-1.2511,+0.2777]\).
The paired intervals for this change and both models' calibration effects
include zero, leaving the allocation benefit of recalibration uncertain.

\paragraph{Cost weighting with action order fixed.}
Keeping each policy's scores and action order fixed, we compare recorded
transition costs (256 reasoning tokens plus the answer probe) with a
diagnostic ledger charging 256 for every transition.  Uniform weighting
changes the DeepSeek-minus-Qwen3 relative-gain gap by \(-0.00101\) percentage
points, with interval \([-0.00210,+0.00001]\).  Answer probes comprise about
1.3\% of transition tokens; this contrast isolates their cost variation.
Signal-acquisition costs are evaluated
separately in \Cref{app:cost-sensitivity}.

\subsection{Answer persistence and local recovery}
\label{app:cpu-persistence}

Persistence \(p\) is the fraction of up to four recent probe answers
matching the current answer.  We group wrong checkpoints by \(p\) and
estimate recovery as the number of correct continuation branches divided by
the total number of branches.  This single-draw rate uses the 256-token
answer-rerun branch labels from the continuation-length ablation.  Replay and
repair retention instead use the primary audit's answer-now labels.
\Cref{tab:cpu-persistence} reports all persistence groups, including missing
values.

\begin{table}[t]
\centering
\small
\setlength{\tabcolsep}{3pt}
\begin{tabular}{lrrl}
\toprule
Persistence & \(N\) & Recovery (\%) & 95\% CI \\
\midrule
\multicolumn{4}{l}{\textbf{Qwen3}} \\
\(p<0.5\) & 290 & 31.29 & \([23.07, 42.22]\) \\
\(0.5\le p<1\) & 497 & 21.03 & \([15.48, 27.89]\) \\
\(p=1\) & 571 & 10.29 & \([6.24, 17.15]\) \\
Missing & 495 & 48.13 & \([41.39, 55.36]\) \\
\midrule
\multicolumn{4}{l}{\textbf{DeepSeek}} \\
\(p<0.5\) & 409 & 27.93 & \([21.34, 36.23]\) \\
\(0.5\le p<1\) & 614 & 18.65 & \([14.30, 23.95]\) \\
\(p=1\) & 811 & 5.43 & \([3.41, 8.27]\) \\
Missing & 412 & 40.78 & \([33.19, 48.80]\) \\
\bottomrule
\end{tabular}
\caption{Single-draw recovery by persistence on the 256-token answer-rerun labels. \(N\) counts wrong checkpoints, each with four branches. Intervals cluster checkpoints and branches by problem.}
\label{tab:cpu-persistence}
\end{table}

Wrong answers with \(p=1\) recover in 10.29\% of Qwen3 branches and 5.43\%
of DeepSeek branches, compared with 31.29\% and 27.93\% when \(p<0.5\).
High persistence is therefore associated with lower local recovery among
wrong answers: stability can accompany errors that resist a short
continuation.

\subsection{Do observed repairs persist?}
\label{app:cpu-retention}

Repair retention measures whether a wrong-to-correct transition is followed
by \(h\) consecutive correct checkpoints on the stored spine.  We use the
same 200-problem subset, with 256 reasoning tokens between checkpoints.
Each horizon includes only repair events with complete follow-up, so sample
sizes differ across horizons.  A problem can contribute multiple events,
which are grouped together in bootstrap resampling.

\begin{table}[t]
\centering
\small
\setlength{\tabcolsep}{3pt}
\begin{tabular}{rrrl}
\toprule
\(h\) & \(N\) & Remain correct (\%) & 95\% CI \\
\midrule
\multicolumn{4}{l}{\textbf{Qwen3}} \\
1 & 486 & 89.92 & \([85.40, 94.41]\) \\
4 & 458 & 83.19 & \([77.33, 88.96]\) \\
\midrule
\multicolumn{4}{l}{\textbf{DeepSeek}} \\
1 & 415 & 85.78 & \([79.86, 91.61]\) \\
4 & 387 & 81.65 & \([75.12, 87.81]\) \\
\bottomrule
\end{tabular}
\caption{Fraction of observed repairs remaining correct at every one of the next \(h\) checkpoints. \(N\) counts repair events with complete follow-up. Each step adds 256 reasoning tokens; intervals cluster events by problem.}
\label{tab:cpu-retention}
\end{table}

Over the next four checkpoints, 83.19\% of eligible Qwen3 repairs and 81.65\%
of DeepSeek repairs remain correct throughout (\Cref{tab:cpu-retention}).
Many observed repairs therefore persist beyond the immediate next answer.

\subsection{Automatic readout diagnostics}
\label{app:cpu-readout}

For each model, a fixed hash selects 100 distinct confirmation problems
with at least one wrong checkpoint, then one wrong checkpoint per problem,
without using continuation outcomes.  Decoding saved answer tokens with the
pinned tokenizers and applying the primary audit's grader reproduces all
200 answer-now labels.

In 6 of the 100 Qwen3 cases and 24 of the 100 DeepSeek cases, the last
complete \texttt{boxed} expression in the reasoning prefix passes the
grader although the answer-now probe fails.  This is prefix--readout
disagreement: a prefix can contain an accepted answer even when the elicited
final answer is incorrect.  The boxed expression may precede later text,
so the diagnostic concerns answer agreement rather than the validity of the
complete reasoning.  It motivates distinguishing reasoning errors from
failures to read out an answer already present in the prefix.

\paragraph{Interpretation and scope.}
Together, these analyses locate allocation gains across datasets and
characterize recovery in stored trajectories.  They describe associations
among grader-defined outcomes: persistence groups can differ in depth and
difficulty, and the readout sample consists of problems with a wrong
checkpoint.  Independent trajectory regeneration and human error annotation
would extend this evidence to generation variability and semantic error
types, supporting investigation of the mechanisms behind recovery.

\end{document}